\documentclass[lettersize,journal]{IEEEtran}
\usepackage{amsmath,amsfonts}
\usepackage{algorithmic}
\usepackage{multirow}
\usepackage[linesnumbered,ruled]{algorithm2e}
\usepackage{array}
\usepackage[caption=false,font=normalsize,labelfont=sf,textfont=sf]{subfig}
\usepackage{textcomp}
\usepackage{stfloats}
\usepackage{url}
\usepackage{verbatim}
\usepackage{graphicx}
\usepackage{cite}
\begin{document}
 
\title{
A Multi-objective Complex Network Pruning Framework Based on Divide-and-conquer and Global Performance Impairment Ranking
}

\author{{
~Ronghua~Shang,~\IEEEmembership{Senior~Member,~IEEE},
~Songling~Zhu$^{*}$,
~Yinan~Wu,
~Weitong~Zhang,~\IEEEmembership{Member,~IEEE},
~Licheng~Jiao,~\IEEEmembership{Fellow,~IEEE},
~Songhua~Xu
\vspace{-7mm}
}

\thanks{
This work was partially supported by the National Natural Science Foundation of China under Grants Nos. 62176200 and 62271374, the Natural Science Basic Research Program of Shaanxi under Grant Nos. 2022JC-45 and 2022JQ-616, the 111 Project, the Guangdong Basic and Applied Basic Research Foundation under Grant No. 2021A1515110686, the Research Project of SongShan Laboratory under Grant YYJC052022004, and the Postdoctoral Research Project Funding of Shaanxi under Grant 2023BSHTBZZ30.
}

\thanks{
Ronghua Shang, Songling Zhu, Yinan Wu, Weitong Zhang, Licheng Jiao are with the Key Laboratory of Intelligent Perception and Image Understanding of Ministry of Education, School of Artificial Intelligence, Xidian University, Xi’an, Shaanxi Province 710071, China (e-mail: rhshang@mail.xidian.edu.cn, slzhu@stu.xidian.edu.cn, ynwu25@stu.xidian. edu.cn, wtzhang\_1@xidian.edu.cn, lchjiao@mail.xidian.edu.cn).
Songhua Xu is with the Department of Health Management \& Institute of Medical Artificial Intelligence, The Second Affiliated Hospital of Xi’an Jiaotong University, Xi’an, China. (e-mail: songhuaxu@mail.xjtu.edu.cn).
}

}


\maketitle

\begin{abstract}
Model compression plays a vital role in the practical deployment of deep neural networks (DNNs), and evolutionary multi-objective (EMO) pruning is an essential tool in balancing the compression rate and performance of the DNNs. However, due to its population-based nature, EMO pruning suffers from the complex optimization space and the resource-intensive structure verification process, especially in complex networks. To this end, a multi-objective complex network pruning framework based on divide-and-conquer and global performance impairment ranking (EMO-DIR) is proposed in this paper. Firstly, a divide-and-conquer EMO network pruning method is proposed, which decomposes the complex task of EMO pruning on the entire network into easier sub-tasks on multiple sub-networks. On the one hand, this decomposition narrows the pruning optimization space and decreases the optimization difficulty; on the other hand, the smaller network structure converges faster, so the proposed algorithm consumes lower computational resources. Secondly, a sub-network training method based on cross-network constraints is designed, which could bridge independent EMO pruning sub-tasks, allowing them to collaborate better and improving the overall performance of the pruned network. Finally, a multiple sub-networks joint pruning method based on EMO is proposed. This method combines the Pareto Fronts from EMO pruning results on multiple sub-networks through global performance impairment ranking to design a joint pruning scheme. The rich experiments on CIFAR-10/100 and ImageNet-100/1k are conducted. The proposed algorithm achieves a comparable performance with the state-of-the-art pruning methods.
\end{abstract}


\section{Introduction}
\IEEEPARstart{W}{ith} the development of the artificial intelligence technology, deep neural networks (DNNs) have been successfully utilized with many applications in areas, such as SAR image processing \cite{shang_hyperspectral_2022}, 
object tracking \cite{wang_dynamic_2021,wu_cslt_2023},
and semantic segmentation \cite{li_ctnet_2022, li_deep_2022}. These successes are due to the increasing ability of feature representation and learning. However, as the performance grows, so do their storage and computational burden. For example, AlexNet has 8 layers of neural networks and 714M floating point operations \cite{krizhevsky_imagenet_2012}, GoogLeNet has 22 layers of neural networks and 1.5G floating point operations \cite{szegedy_going_2015}, while VGG16 has 16 layers of neural networks and 15.5G floating point operations \cite{simonyan_very_2014}. 
They are computationally expensive, especially for embedded devices and mobile terminals, which have difficulty obtaining adequate computing resources \cite{han_learning_2015}. 
To address this problem, various neural network compression methods have been proposed, 
such as knowledge distillation 
\cite{hinton_distilling_2015, zhang_student_2022},
network pruning 
\cite{yang_skeleton_2023,feng_automatically_2022},
and model quantization \cite{lin_siman_2022, cai_zeroq_2020}. Among them, network pruning has attracted extensive attention from researchers due to its excellent compression performance \cite{xu_convolutional_2020}.

Existing pruning algorithms fall into two major categories. 
The first is unstructured pruning, which obtains a compressed neural network structure by discovering and pruning unimportant weights.
Han et al. \cite{han_learning_2015} considered that the weights below a certain threshold are unimportant and can be pruned. 
Guo et al.  \cite{guo_dynamic_2016} designed a pruning algorithm with the dynamic method, which enables the restoration of the connections when the pruned weight is found to be significant. Unstructured pruning performs well, but it usually demands special hardware for acceleration \cite{lin_filter_2021}. The other is structured pruning, which prunes network layers or filters instead of the network weights. In contrast to unstructured pruning, the pruned network maintains the conventional convolutional kernel structure and can be accelerated for training and inference using standard deep-learning libraries. 
He et al.  \cite{he_soft_2018} developed a soft filter pruning, enabling these pruned structures to be updated during the following training. 
Ayinde et al.  \cite{ayinde_building_2018} identified redundant network feature structures by the relative cosine distance. 
Yu et al.  \cite{yu_nisp_2018} performed pruning by the importance measured for each neuron in the final response layer. 
He et al. \cite{he_filter_2019} used the geometric median to find the most replaceable filter. 
Lin et al. \cite{lin_towards_2019} designed sparse structures by generating adversarial learning. 
Lin et al. \cite{lin_hrank_2020} found that lower-rank features contained less important information and had less impact on the network. 
He et al. \cite{he_asymptotic_2020} designed a progressively increasing compression rate to gradually concentrate the information learned from the dataset into the remaining filters. 
Zhang et al. \cite{zhang_filter_2021} pruned from the frequency domain. 
Chen et al. \cite{chen_dynamical_2021} performed pruning by the conditional accuracy change of each channel under channel-wise gate control. 
He et al. \cite{he_learning_2020} devised a learnable and optimizable method to choose suitable pruning criteria for different network layers.
Li et al. \cite{li_pruning_2017} proposed a filter-based pruning algorithm that determines the importance of a filter to the network by its norm. 
However, the relationship between the different structures in a neural network is very complex \cite{yosinski_how_2014}, and the absence of any part of the network structure can impact the performance of the entire network. Therefore, whether structured pruning or unstructured pruning, these statistical information-based measures of network redundancy may not reflect the network performance  after pruning accurately \cite{niu_exploiting_2022}.

The evolutionary algorithm is a bio-inspired optimization algorithm based on individual fitness that can accurately measure the solution's performance and solve complex optimization tasks \cite{zhao_two-stage_2022}. Its powerful optimization capabilities have attracted attention in different fields. 
Shang et al. \cite{shang_evolutionary_2022} designed a complex network architecture search algorithm with a genetic algorithm. 
Stodola et al. solved multi-depot vehicle routing problems with an ant colony algorithm \cite{stodola_adaptive_2022}. 
With the help of the PSO algorithm, Huang et al. \cite{huang_particle_2022} designed a compact network structure search algorithm. In addition, there have been several successful attempts to use evolutionary algorithms to optimize complex pruning problems on neural networks. 
Zhou et al. \cite{zhou_evolutionary_2020} proposed an evolutionary multi-objective (EMO) neural network compression method ECDNN, which finds suitable pruned network structures for biomedical image segmentation. 
Zhou et al. \cite{zhou_knee-guided_2021} proposed a knee-point guided evolutionary multi-objective algorithm (KGEA), utilizing an evolutionary multi-objective algorithm to find a pruning structure, which could make a trade-off between the parameter scale and pruning performance. However, the pruning space of deep neural networks is very complex, making their optimization by evolutionary multi-objective algorithms extremely challenging. 
Zhou et al. \cite{zhou_evolutionary_2021} designed a network block-based multi-objective network pruning algorithm, which prunes the network block containing multiple layers instead of the network layers. Although this approach reduces the pruning space of complex networks to some extent, the optimization space still remains challenging. The network block-based pruning cannot optimize the redundancy of the filters within the block, which constrains the algorithm's performance. In addition, pruned networks are intensive in computation and storage. Their training and performance verification consume a large number of computational resources. Therefore, to better exploit the performance of evolutionary multi-objective algorithms, the optimization space and the resource consumption of the network pruning task should be reduced.

Divide-and-conquer \cite{bentley_multidimensional_1980} is an effective way for complex tasks, which decomposes a complex task into several simpler sub-tasks, thus reducing the complexity of the whole task. This approach has been used in a variety of research areas. 
Gidiotis et al. \cite{gidiotis_divide-and-conquer_2020} decomposed the difficult summarization task of the whole document into simpler sub-tasks of several small parts, improving the summarization performance.
Xiong et al. \cite{xiong_open_2019} found that the difficulty of visual counting increases with the object's density, and reducing the density in the recognition region by the divide and conquer method can improve the performance of object counting.
Compared to the complex optimization task, algorithms focusing on simple sub-tasks perform better. Therefore, the divide-and-conquer approach may be suitable for the complex multi-objective network structure pruning problem.
However, the network is an entire one made up of multiple, closely interconnected parts, and the change in any part has a non-negligible effect on the entire.
If a sub-network is optimized and trained separately, it may not collaborate well with the other sub-network from the same network, affecting the overall performance. In addition, it is an open question of how to design a complete pruning scheme using these multi-objective optimization results from multiple sub-networks optimized independently. Therefore, it is impractical to apply the idea of divide-and-conquer directly to the pruning task on the complex network structure.

Inspired by the aforementioned questions, this paper proposes 
a multi-objective complex network pruning framework based on divide-and-conquer and global performance impairment ranking (EMO-DIR),
which aims to reduce the optimization space and the resource-consuming of pruning complex networks while making independently optimized sub-network pruning tasks cooperate well to achieve better overall performance.
Firstly, a divide-and-conquer EMO network pruning method is proposed, which decomposes the complex multi-objective pruning task on the complete network into the simple pruning task on multiple sub-networks. The optimization complexity of the complete network is the product of the complexities of the several sub-networks. In contrast, the total optimization complexity of multiple sub-networks is the sum of their complexities. Therefore, our method is computationally efficient. In addition, the sub-networks have fewer parameters and converge faster than the complete network structure. Thus, their training consumes fewer computation resources. Secondly, a sub-network training method based on cross-network constraints is designed to improve the collaboration among sub-networks optimized independently. Different parts of the network process different features, and the output features from the previous sub-network are used as input features for the following one. The proposed sub-network training method ensures that the next sub-network can process the features generated by the previous one well by constraining the output features from the previous one and the input features of the following one, making the independently pruned sub-networks collaborate well and improve the overall performance of the pruned network.
Finally, this paper proposes a multiple sub-networks joint pruning method based on EMO, which calculates a global performance impairment index based on the Pareto Front obtained from multiple sub-network optimization tasks. This index reflect the impact of the sub-network pruning scheme on the whole network's performance, thus helping the algorithm design a joint pruning scheme. Extensive extensive experiments are performed on the CIFAR10, CIFAR100, ImageNet-100 and ImageNet-1k datasets, and the experimental results demonstrate the effectiveness and efficiency of the proposed algorithm. The contributions are as follows.

\begin{itemize}
\item A divide-and-conquer EMO network pruning method is proposed. It decomposes the multi-objective pruning task on a complex network into simple pruning tasks on multiple sub-networks. This method reduces the optimization difficulty of the multi-objective pruning algorithm and the resource consumption of validating the pruning structures during the evolutionary process.

\item A sub-network training method based on cross-network constraints is designed. It constrains the output features of the previous sub-network and the input features of the next one, so the output features of the previous sub-network can be processed by the next one, thus improving the overall network performance.

\item A multiple sub-networks joint pruning method based on EMO is proposed. It calculates the global performance impairment index based on the non-dominated solutions from the evolutionary multi-objective pruning algorithm on each sub-network. The global performance impairment index reflects the impact of different pruning schemes of the sub-networks on the whole network's performance. Thus, this index can help the algorithm design a joint pruning scheme.
\end{itemize}

In the following, this paper first presents the details of the proposed algorithm; then, it describes the related experimental setup, experimental results, and analyses; finally, the conclusion and future work are presented.

\section{The proposed algorithm}
\begin{figure*}[!b]
\centering
\includegraphics[scale=0.405]{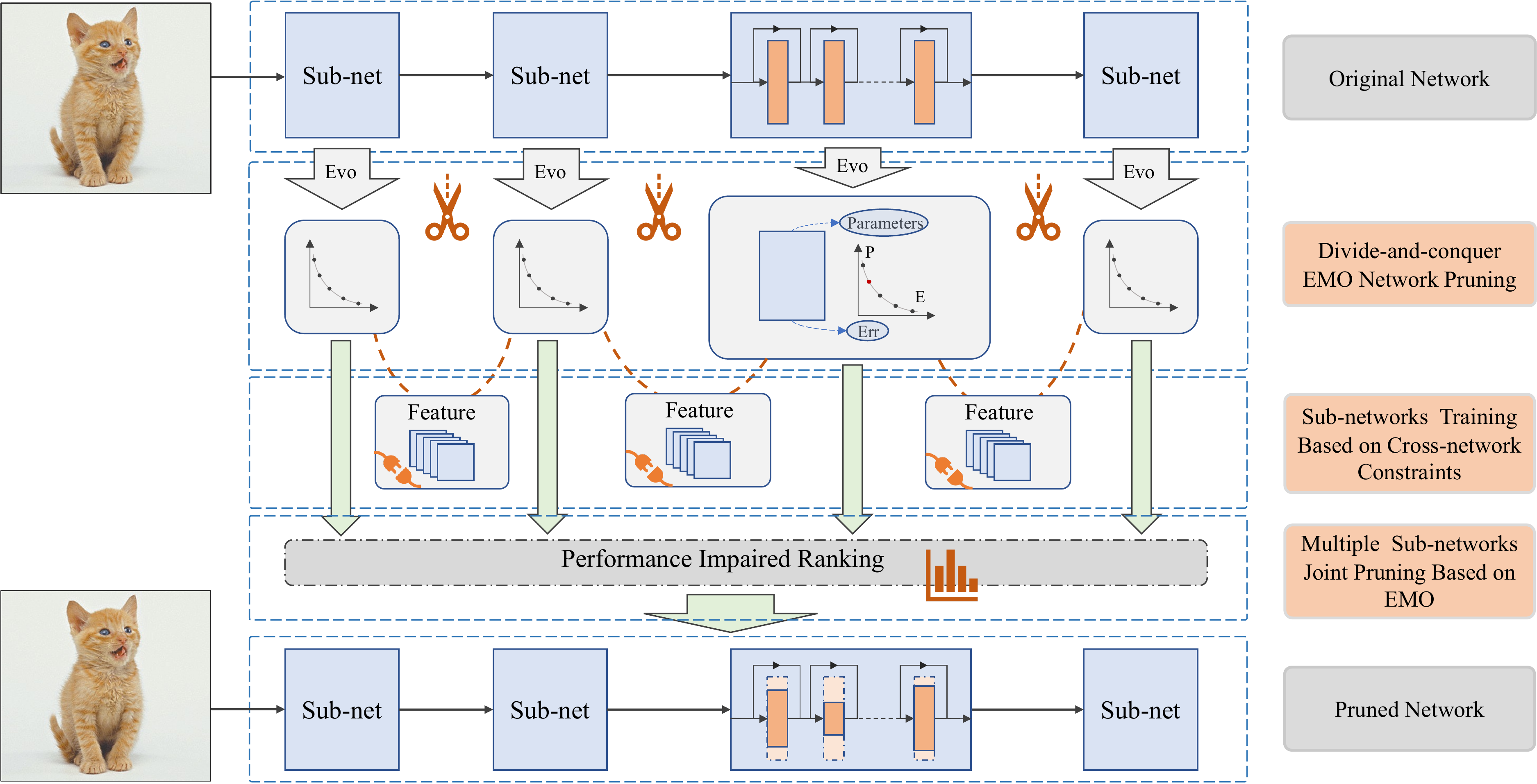}
\caption{The overall structure of 
	the EMO-DIR.
	}
\label{fig1}
\end{figure*}

Evolutionary multi-objectives based network pruning algorithms drive a trade-off between the performance and resource cost of the network to generate a better network pruning structure. However, the population-based nature makes these algorithms suffer from the complex optimization space and the highly resource-consuming individual validation process. To this end, this paper proposes 
a multi-objective complex network pruning framework based on divide-and-conquer and global performance impairment ranking (EMO-DIR).
It decomposes the complex network pruning task into multiple simple sub-network pruning tasks, thereby narrowing the pruning optimization space of the EMO algorithm and the resource consumption of the individual performance validation. Fig. \ref{fig1}  describes the overall structure of the EMO-DIR.

Fig. \ref{fig1} shows that the algorithm achieves complex EMO pruning on the original network by several multi-objective pruning on sub-networks.
First, this paper designs a divide-and-conquer EMO network pruning method, which performs independent EMO pruning on each sub-network to finish the whole pruning. It considers the sub-network parameter number and prediction error as two optimization objects.
During the training of the sub-networks, to improve the collaboration among the different sub-networks, this paper proposes a sub-networks training method based on cross-network constraints, which constrains the features between adjacent sub-networks and helps the next sub-network to process the output features from the previous one.
Finally, to generate a whole pruning scheme with the non-dominated solutions produced by the EMO pruning on these sub-networks, we design a multiple sub-networks joint pruning method based on EMO. With the help of the global performance impairment ranking, all the sub-network optimization results are combined to produce a complete pruning scheme.


\begin{figure*}[h]
	\includegraphics[scale=0.46]{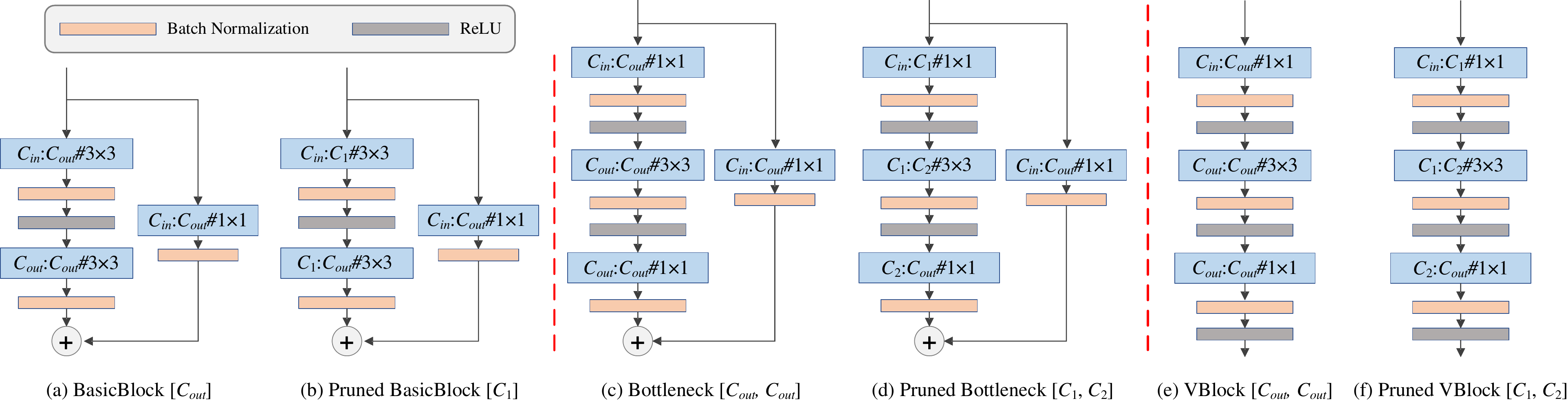}
	\caption{The codings of the original network structures and their pruned structures (BasicBlock, Bottleneck residual structures and VGG block VBlock).}
	\label{fig2}
\end{figure*}

\subsection{Encoding and Decoding of Network Structures}
The evolutionary algorithm cannot directly optimize the network structure but needs to transform the network into structure coding. The coding optimized by the evolutionary algorithm must also be transformed into pruned network structures for training and inference. Three network structures are involved in the experiments: BasicBlock, Bottleneck residual structures, and VGG Block. Fig. \ref{fig2} describes the corresponding structure codings of the original and the pruned. These codings are marked by ``[ ]", where the output channel number is used as the coding content of the convolutional layer. ``$C_{in}:C_{out}\#3\times3$" means that the input channel number from the convolution kernel is $C_{in}$, the output number is $C_{out}$, and the size of the convolution kernel is $3\times3$.

%
%


As shown in Fig. \ref{fig2}, for residual structures, the output channel number in the last layer of the residual block remains unchanged during pruning. 
For VGG structures, we consider all network layers in the sub-network as one block and keep the output channel of the last one unchanged. Fig. \ref{fig2} (c) and (d) use three layers for clear demonstration, and one block structure can contain more.
Because this method keeps the shape of the input and output features unchanged, it does not require any additional adjustment when connecting these pruned sub-network structures into a complete network, thus simplifying the pruning process. Only the output channel number except the last layer is encoded when encoding these network structures. For example, in Fig. \ref{fig2}(b), the pruning structure for the BasicBlock with two convolutional layers is encoded as $[C_1]$. In Fig. \ref{fig2}(d), the pruning structure for the Bottleneck with three convolutional layers is encoded as $[C_1, C_2]$. The coding is a sequential stacking of internal network structures. For example, for a residual sub-network coding $[[C_{11}, C_{12}], [C_{21}, C_{22}], \cdots, [C_{n1}, C_{n2}]]$, the internal structure is a sequentially connected structure of three BasicBlock residual structures encoded as $[C_{11}, C_{12}]$, $[C_{21}, C_{22}]$ and $[C_{n1}, C_{n2}]$. For a VGG sub-network coding $[C_{1}, C_{2}, \cdots, C_{n}]$, the internal structure is n+1 network layers connected sequentially.

\subsection{
The Divide-and-conquer EMO Network Pruning Method
}
The divide-and-conquer EMO network pruning method
aims to solve the EMO pruning problem for complex network structures using the divide-and-conquer idea. For the convenience of description, this paper uses Equation (\ref{eq1}) to define the ordered set of complete networks.

\vspace{-2mm}
\begin{equation}
\label{eq1}
Net_{all} =\{l_1, l_2, \cdots, l_N\}
\end{equation}
where $N$ means the total number of network layers, and each network consists of multiple ordered network layers.

Each neural network layer performs simple feature transformation tasks, while a group of neural network layers can perform a more complex feature processing task. Thus, as shown in Equation (\ref{eq2}), a complete network can be considered a set of multiple feature-processing sub-networks.

\begin{equation}
\label{eq2}
\begin{aligned}
X_{out} &= F(Net_{all}, X_{in}) \\
&= F(l_N, F(l_{N-1}, \cdots F(l_1, X_{in}))) \\
&= F(SubNet_M, F(SubNet_{M-1}, \\
& \qquad \cdots F(SubNet_1, X_{in})))
\end{aligned}
\end{equation}
where $F(Net, X)$ is used to describe the feature processing process of the network with the ordered set $Net$ and the features to be processed $X$. $X_{out}$ is the network's final output. Equation (\ref{eq2}) shows that the original network structure $Net_{all}$ can be divided into several sub-networks $\{SubNet_1, SubNet_2,\cdots, SubNet_M\}$, and each sub-network structure completes a part of the feature processing task. $M$ denotes the number of sub-network structures, i.e., the number of sub-tasks into which the whole feature processing task is divided. The set of sub-networks consists of several ordered network layers that do not overlap, as shown in Equation (\ref{eq3}).

\vspace{-2mm}
\begin{equation}
\label{eq3}
\left\{\begin{array}{l}
SubNet_1 \cup SubNet_2 \cup \cdots \cup SubNet_N=Net_{all} \\
SubNet_1 \cap SubNet_2 \cap \cdots \cap SubNet_N=\emptyset 
\end{array}\right.
\end{equation}

From Equation (\ref{eq3}), the complete network $Net_{all}$ comprises several sub-networks. From Equation (\ref{eq2}), the feature processing function of the complete network can also be implemented by multiple sub-networks $SubNet_i$. Thus, a pruning task on a complete network can be decomposed into multiple pruning tasks on sub-networks. Equation (\ref{eq4}) shows the multi-objective pruning task on the complete network.

\vspace{-2mm}
\begin{equation}
\label{eq4}
\left\{\begin{aligned}
\underset{Net_{all}^{\prime}}{\arg \min} & \left(P\left(Net_{all}^{\prime}\right), E\left(Net_{all}^{\prime}, X, Y\right)\right) \\
s.t. \quad & P\left(li^{\prime}\right) \leq P(li), \forall li \in Net_{all}, \forall li^{\prime} \in Net_{all}^{\prime}
\end{aligned}\right.
\end{equation}

The optimization objectives of the multi-objective pruning task are the parameter number $P(Net_{all}^{\prime})$ and the prediction error $E(Net_{all}^{\prime}, X, Y)$, where $X$ and $Y$ mean the datasets and labels to be processed. The optimization algorithm could find the pruned network structure $Net_{all}^{\prime}$ by pruning the original network $Net_{all}$. $Net_{all}^{\prime}$ performs well on these two optimization objectives.
The pruning space of the original network $Net_{all}$ is very large. For a neural network with $N$ network layers and the $i$-th network layer has $k_i$ convolutional kernels, the size of its pruning optimization space $O(Net_{all})$ is shown in Equation (\ref{eq5}).

\vspace{-0mm}
\begin{equation}
\label{eq5}
O\left(Net_{all}\right)=\prod_{i=1}^N O\left(k_i\right)
\end{equation}

If the network with $N$ layers is divided into $M$ sub-networks, the multi-objective pruning task represented by Equation (\ref{eq4}) can be divided into pruning tasks of the $M$ sub-networks shown in Equation (\ref{eq6}).

\vspace{-2mm}
\begin{equation}
\label{eq6}
\left\{\begin{aligned}
\underset{Net_i^{\prime}}{\arg \min } & \left(P\left(Net_i^{\prime}\right), E\left(N_{e t_i^{\prime}}, X, Y\right)\right) \\
s.t. \quad & P\left(li^{\prime}\right) \leq P(li), \forall li \in Net_i, \forall li^{\prime} \in Net_i^{\prime}
\end{aligned}\right.
i \in [1,M]
\end{equation}

The size of the pruning space $O'(Net_{all})$ is shown in Equation (\ref{eq7}) and is significantly smaller than the pruning space $O(Net_{all})$ from the original network.

\vspace{-1mm}
\begin{equation}
\label{eq7}
\begin{aligned}
O^{\prime}\left(N e t_{all}\right)
&=\sum_{j=1}^M O\left(N e t_j\right) \\
&=\sum_{j=1}^M \prod_{i=1}^{N_j} O\left(k_{i+N_1+N_2+\cdots N_{j-1}}\right) \\
&<<\prod_{i=1}^N O\left(k_i\right) =O\left(Net_{all}\right)
\end{aligned}
\end{equation}
where $N_j$ denotes the number of network layers in the $j$th sub-network. Therefore, dividing the original complex network into several simpler sub-networks significantly reduces the optimization space of the evolutionary multi-objective network pruning algorithm, thus reducing the optimization difficulty. Furthermore, Equation (\ref{eq2}) describes that the feature processing of the whole network can be regarded as a set of ordered sub-network processing. Therefore, if the pruned sub-network retains the feature processing capability of the original sub-network, or if its feature processing capability is only slightly impaired, the impact of pruning on the original network will also be minor.

\begin{figure}[h]
\centering
\includegraphics[scale=0.42]{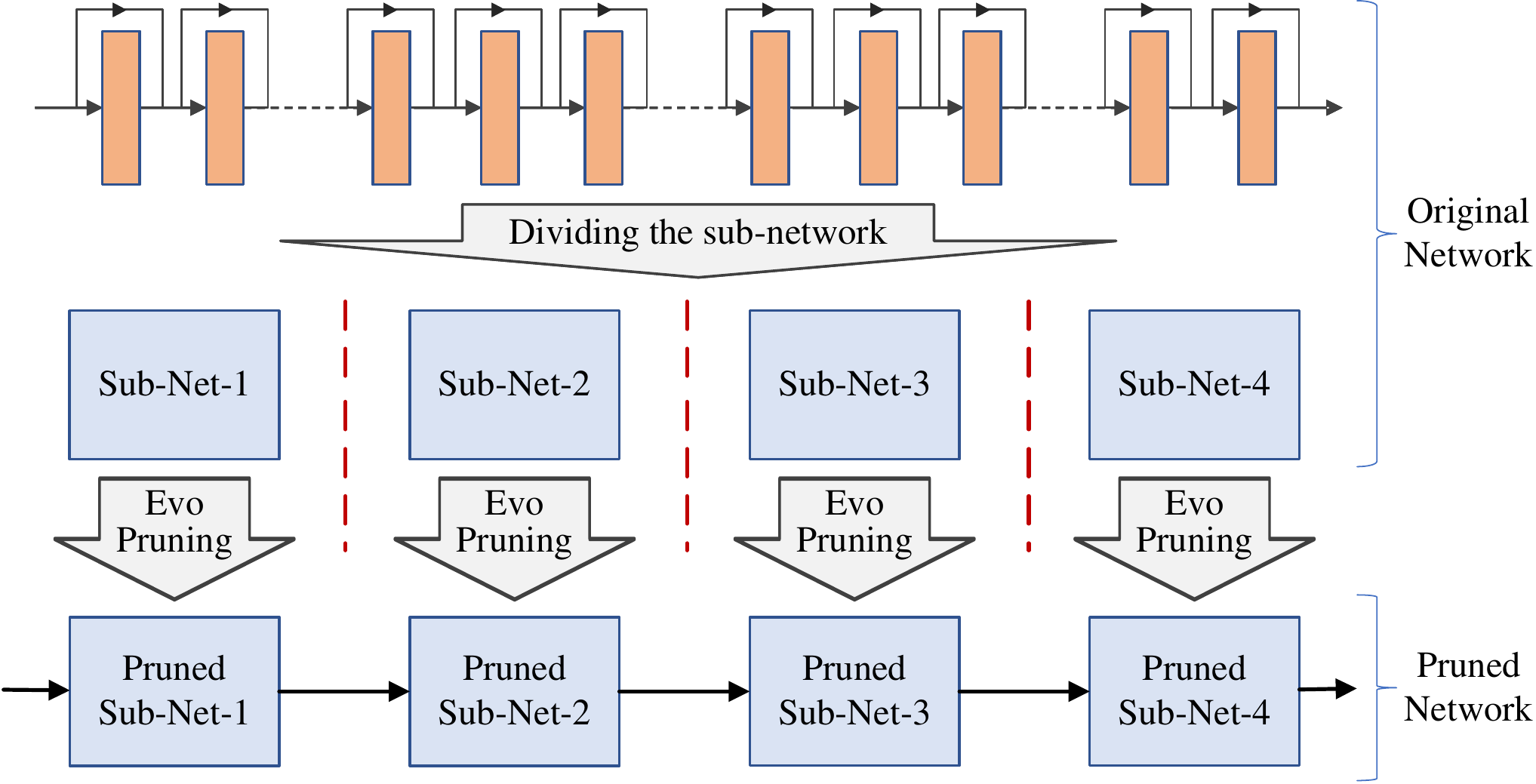}
\caption{
The divide-and-conquer EMO network pruning method.
}
\label{fig3}
\end{figure}

Inspired by this, we design
a divide-and-conquer EMO network pruning method. The major structure is shown in Fig. \ref{fig3}.

As shown in Fig. \ref{fig3}, this network pruning method first divides the whole neural network, which will be pruned, into multiple sub-networks that do not overlap. Each sub-network consists of multiple network layers with simple feature processing capability. Then, this method performs an independent evolutionary multi-objective network pruning for each sub-network. The optimization objectives of the evolutionary multi-objectives optimization are the parameter number and the feature processing capability of the sub-network. The EMO algorithm can balance the parameter number and performance of the sub-network to obtain a more significant sub-network compression rate with a minor performance loss. Finally, the algorithm constructs a final pruned network based on the evolutionary multi-objective pruning structure obtained on each sub-network.

The method transforms evolutionary multi-objective pruning from the complete neural network to multiple sub-networks. However, without other help, this method does not accomplish the neural network pruning task well. Firstly, the feature processing capabilities of the sub-networks may change during the independent optimization process. These changes may result in the features they produce not being correctly understood by the following sub-network, thus impairing the network’s overall performance. Secondly, the evolutionary multi-objective pruning in each sub-network is performed independently. The resulting optimization results can only be used to set the pruning scheme for that sub-network rather than for the whole network. To tackle these two problems above, we introduce a sub-network training method based on cross-network constraints and a multiple sub-networks joint pruning method based on EMO, which would be detailed in the following two subsections.

\begin{figure*}[h]
	\centering
	\includegraphics[scale=0.45]{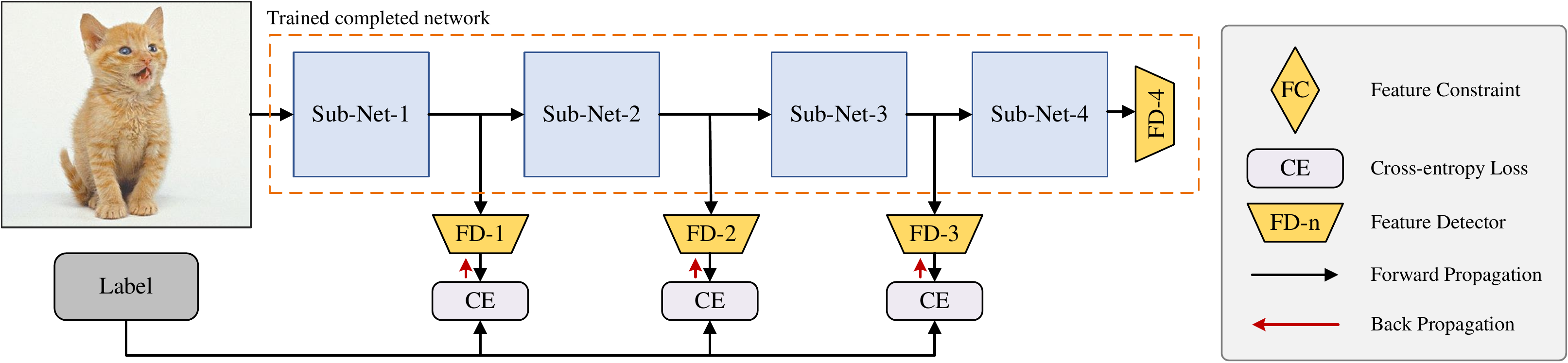}
	
	\vspace{2mm}
	\footnotesize{(a). The training process of feature detectors.}
	\vspace{3mm}
	
	\includegraphics[scale=0.45]{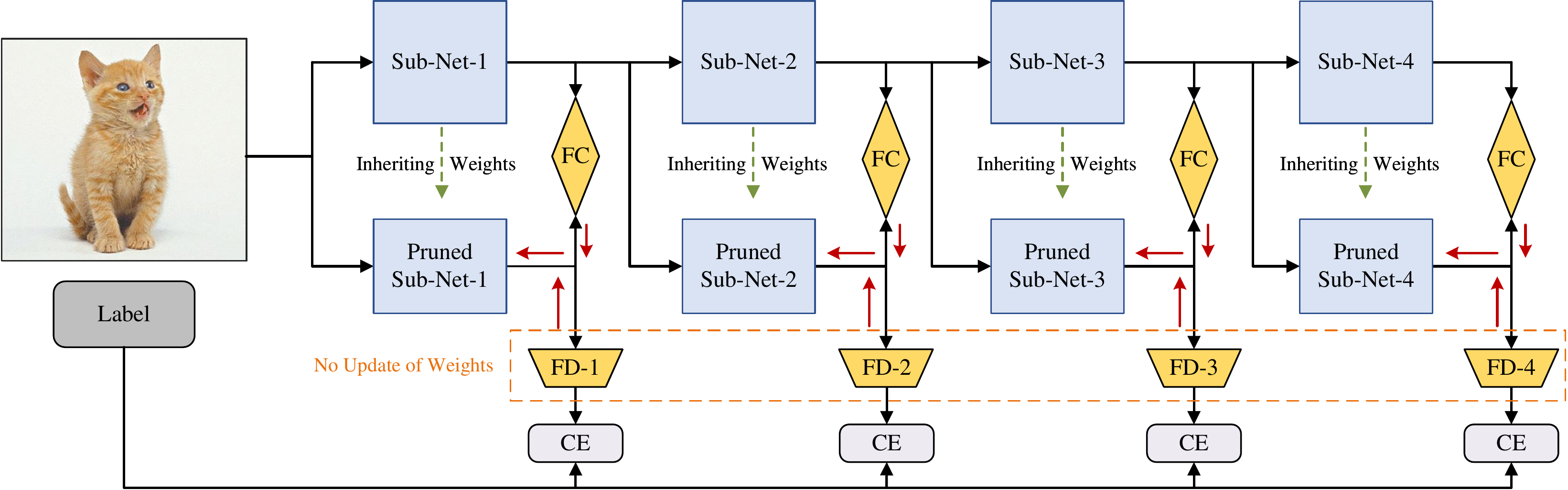}
	
	\vspace{2mm}
	\footnotesize{(b). The training process of multiple sub-networks under feature constraints.}
	
	\caption{The sub-network training method based on cross-network constraints.}
	\label{fig4}
\end{figure*}

\subsection{The Sub-network Training Method Based on Cross-network Constraints}
Multiple sub-networks trained independently destroy the collaborative ability among the sub-networks from the original network, resulting in the next sub-network not being able to process the features generated by the previous one well. This paper proposes a sub-network training method based on cross-network constraints, which improves the collaboration among sub-networks by constraining the previous sub-network output features and the following one's input features. With these constraints, the next one can process the features generated by the previous one well.

The constraint is an element-level feature constraint that is used to shrink the element-level distance between the feature generated by the previous sub-network and the feature that the next sub-network can process. Since the input feature from the pruned network $SubNet_{i+1}^{\prime}$ is the output feature from the original network $SubNet_i$, shrinking the distance between the output features of $SubNet_i^{\prime}$ and $SubNet_i$ can shrink the distance between the input features of $SubNet_{i+1}^{\prime}$ and the output features of $SubNet_i^{\prime}$. As shown in Equation (\ref{eq8}), this paper uses the $L_2$ norm to measure the distance between them.

\vspace{-2mm}
\begin{equation}
\label{eq8}
\begin{aligned}
&Loss_{FC}(SubNet_i, SubNet_{i}^{\prime})=\\
& \quad ||F(SubNet_1+\cdots +SubNet_i, X) - F(SubNet_{i}^{\prime}, \\
& \qquad \qquad \qquad F(SubNet_1+ \cdots + SubNet_{i-1}, X))||_2
\end{aligned}
\end{equation}

The first feature is derived from the output features of the original network at $SubNet_i$, whose input is the dataset $X$. The second feature is derived from the output feature of the pruned network $SubNet_i^{\prime}$, whose input feature is from the original sub-network $SubNet_{i-1}$.

In addition, the feature detector trained with the original network also provides some feature constraints, which are used to encourage the pruned sub-network to perform similar feature processing to the original network. The sub-network does not give equal importance to every element in the feature from the previous sub-network. In features, the complex relationship among elements is also important. Therefore, this paper uses a feature detector to provide an additional feature constraint on the relationship among feature elements. The constraint is implemented through a pre-trained feature detector, which finds specific elemental relationships in the features.
In addition, it can provide performance metrics for sub-network.
Fig. \ref{fig4} describes the proposed method.

\begin{figure}[h]
\centering
\includegraphics[scale=0.45]{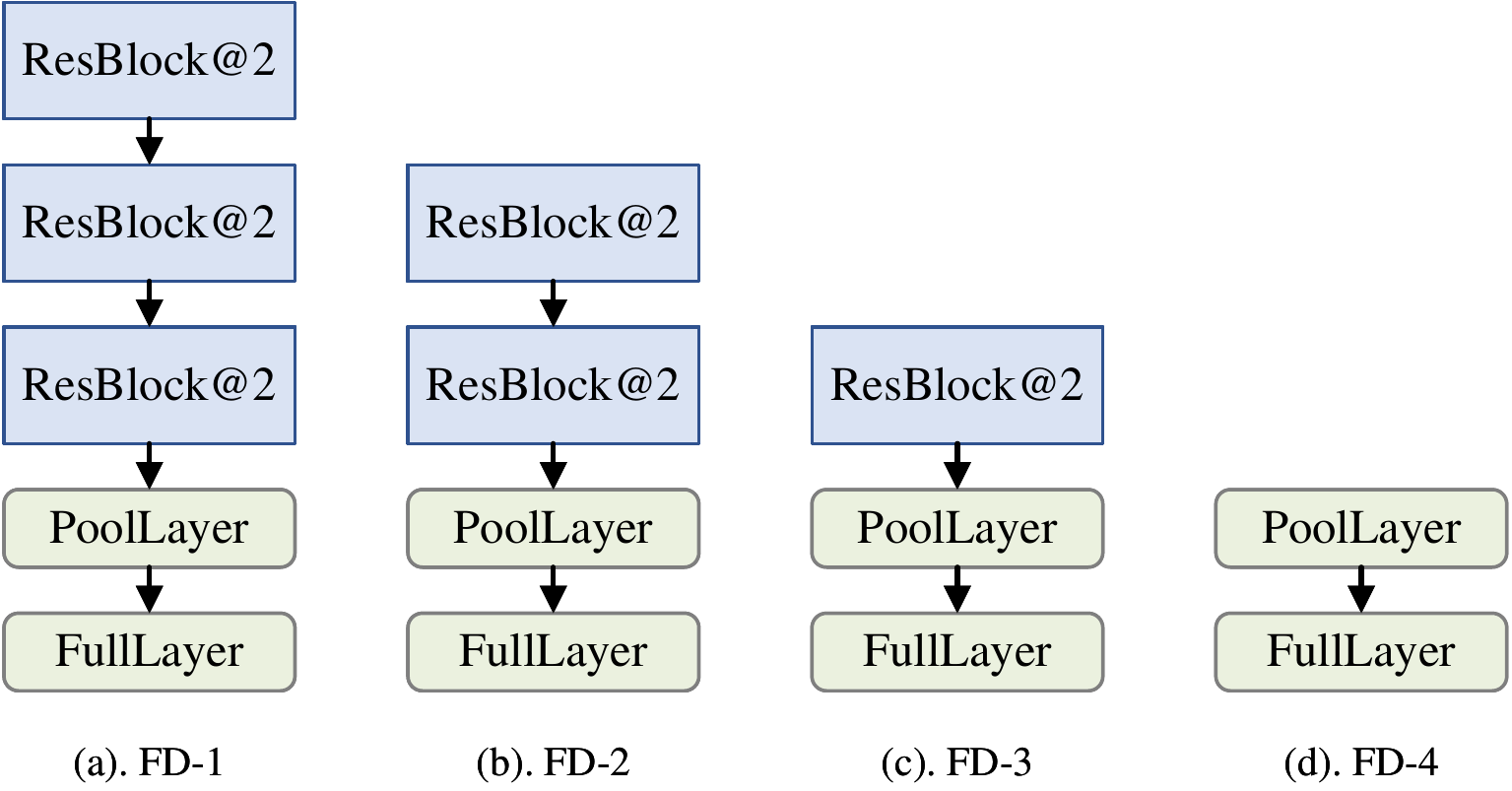}
\caption{The structures of feature detectors located at different positions.}
\label{fig5}
\end{figure}

Fig. \ref{fig4}(a) shows the training process of the feature detectors, and the trained feature detector can identify the element relationships in the features and find the change in the feature processing ability of the sub-network. The input of the feature detector is the intermediate features generated by the sub-network, and the output is the prediction classification. This paper uses the same residual structure to construct the feature detectors. Because the features' spatial resolution and processing difficulty vary, feature detectors on different network locations are different. Fig. \ref{fig5} describes the feature detectors used on the network divided into four sub-networks. 
The feature detectors at different locations contain different numbers of residual blocks. The closer to the output layer from the network, the smaller the number of residual blocks. These residual blocks have the same structure, and they all reduce the spatial resolution of the input features by half. Each feature detector is followed by a pooling and fully connected layer with the same structure as the one used in the original network.
In addition, during the training process, this paper only updates the feature detectors. So, it converges faster and does not impose much computational burden.

When the training processes of the feature detectors are finished,  the training of multiple sub-networks can be followed. This processing is shown in Fig. \ref{fig4}(b). First, the trained feature detectors are connected to the corresponding sub-networks, and their weights are fixed so that the feature detectors do not change during the training process and get stable detection results. 
In addition, the pruned sub-networks randomly inherit some weights from the trained completed network to accelerate model convergence.
Then, the data is fed into the original network to generate input and output features for each pruned sub-network. Finally, the pruned sub-network completes its training process.
During the training of the neural network, loading the training data from the disk is very time-consuming. To save the training time of the sub-networks, this paper generates input and output features for all the sub-networks in one data loading process, then uses these features to complete the forward and backward propagation of all the sub-networks.

The loss function for each sub-network training consists of two components, as shown in Equation (\ref{eq9}). One part is from the element value from the features, which is measured using the $L_2$ norm; the other part is from the relationship among the feature elements, which is measured using cross-entropy.

\vspace{-3mm}
\begin{equation}
\label{eq9}
Loss_i = \frac{1}{2} * (||F_i^{\prime} - F_i||_2 + Loss_{CE}(A_i, Y))
\end{equation}

The Equation (\ref{eq9}) describes the total loss of the $i$th sub-network training.
$F_i$ and $F_i^{\prime}$ denote the output features produced by the $i$th sub-network in the original and pruned networks. $A_i$ denotes the output result of the $i$th feature detector, and $Y$ denotes the ground truth label. $Loss_{CE}(\cdot)$ denotes the cross-entropy loss function.

\subsection{
The Multiple Sub-networks Joint Pruning Method Based on EMO
}
\begin{figure}[h]
\centering
\includegraphics[scale=0.43]{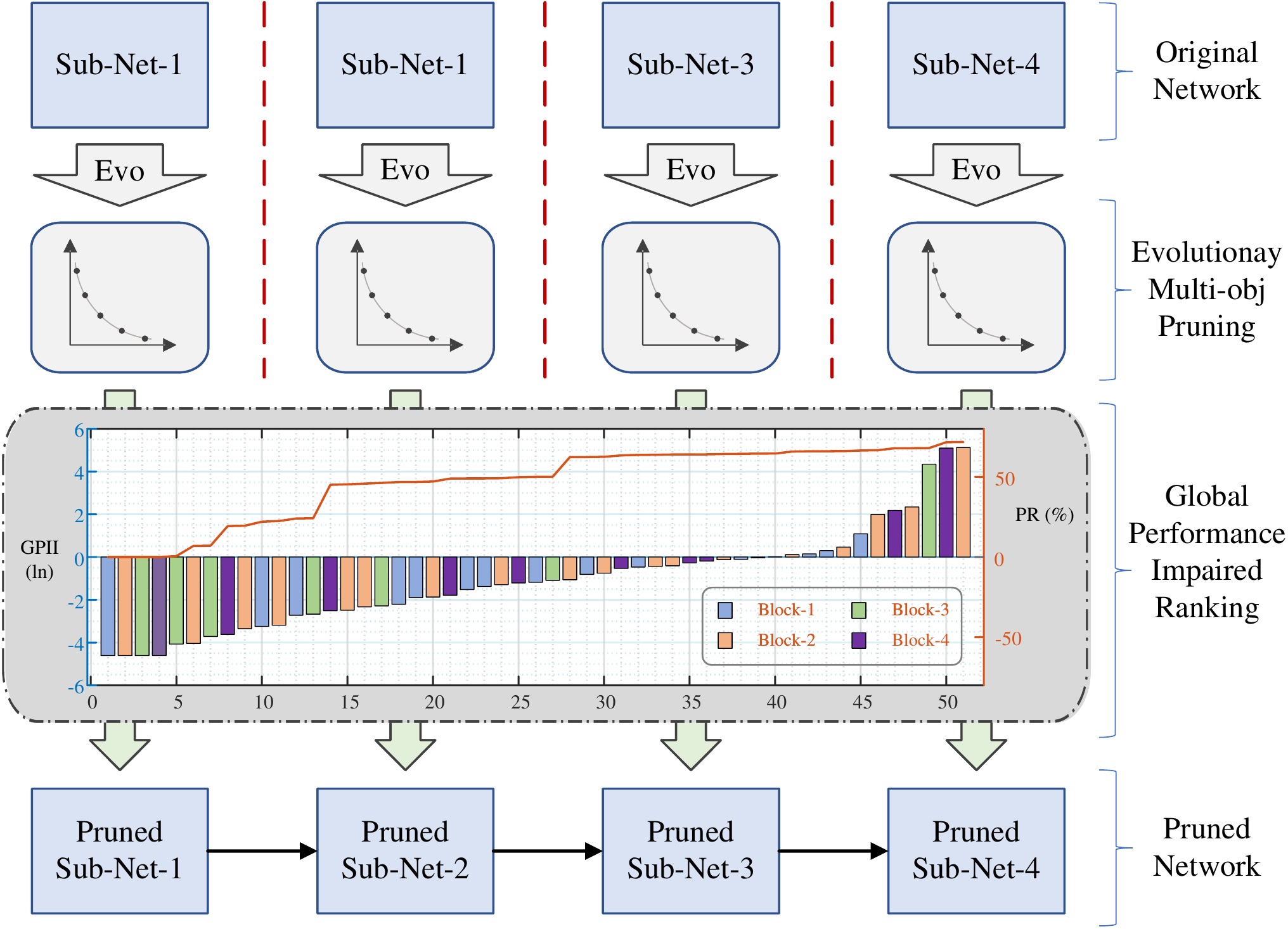}
\caption{
The multiple sub-networks joint pruning method based on EMO.
}
\label{fig6}
\end{figure}

Multiple sub-networks divided from the original neural network can be independently pruned with the EMO optimization algorithm. However, the final pruned neural network structure cannot be directly constructed based on the multi-objective network pruning results on these sub-networks. To this end, this paper proposes 
a multiple sub-networks joint pruning method based on EMO,
which designs a joint pruning scheme based on the multi-objective pruning results on multiple sub-networks. 
Fig. \ref{fig6} illustrates the structure.

Firstly, the algorithm finds the Pareto Front consisting of several non-dominated solutions \cite{deb_multi-objective_2011} from the EMO pruning. These non-dominated solutions perform well in all optimized objectives. The evolutionary multi-objective algorithm did not find any other solution that outperformed them on all optimization objective functions during the optimization process. Then, based on the solutions from the Pareto Front, this paper develops a global performance impairment ranking (GPIR) method that describes what pruning scheme should be adopted for which sub-network has the most negligible expected impact on the network performance.

To obtain GPIR, as shown in Equation (\ref{eq10}), the changes in the parameter number and prediction error relative to the unpruned network are first calculated for each solution.

\begin{equation}
\label{eq10}
\left\{\begin{aligned}
\Delta P_{i, j} = \frac{P_{i, j}-P_{i, baseline}}{P_{i, baseline}} \\
\Delta E_{i, j} = \frac{E_{i, j} - E_{i, baseline}}{E_{i, baseline}}
\end{aligned}
\right.
\end{equation}
where $\Delta P_{i, j}$ and $\Delta E_{i, j}$ represent the change in parameter number and the one in prediction error of the $j$th non-dominated solution on the $i$th sub-network. $P_{i, baseline}$ and $E_{i, baseline}$ are parameter number and prediction error of the unpruned structure on the $i$th sub-network. Next, the global performance impairment index (GPII) can be calculated from $\Delta P_{i, j}$ and $\Delta E_{i, j}$, as shown in Equation (\ref{eq11}).

\begin{equation}
\label{eq11}
I_{i, j} = \frac{\Delta P_{i, j} - min(\Delta P_i)}{\Delta E_{i, j} - min(\Delta E_i) + 0.001}
\end{equation}

The GPII describes the increase in the parameter number caused by per unit increase in error. The smaller the performance impairment index, the better the pruning strategy, and vice versa. It is a measure of the network pruning cost and can also be used to compare the multi-objective pruning results among different sub-networks. This paper uses the performance impairment index to rank the global performance impairment caused by the pruning scheme on different sub-networks to design the joint pruning scheme.

The GPIR between the four sub-networks is shown in Fig. \ref{fig6}, and the corresponding network pruning rate (PR) can be obtained from the ranking result. The pruning rate is also called the compression rate, and this paper calls it the pruning rate for convenience. Fig. \ref{fig6} indicates that the pruning rate increases as the performance impairment index increases. This also demonstrates the flexibility of the proposed algorithm in choosing the pruning rate. Once the multi-objective pruning tasks are completed on all sub-networks, multiple pruning rate design schemes can be obtained. The user can flexibly change the pruning scheme according to the actual requirements without additional repeating the pruning algorithm. For the given pruning rate, the user can find the pruning scheme on each sub-network based on the GPIR, thus determining the global pruning scheme on the original network.

A more detailed procedure of the multiple sub-networks joint pruning method based on evolutionary multi-objective is shown in Algorithm \ref{algorithm1}. First, this method performs independent evolutionary multi-objective pruning on each sub-network and obtains a collection of non-dominated solutions on each sub-network pruning task. These non-dominated solutions from all sub-networks are collected to form the non-dominated set $P_{all}$. Then, the algorithm uses the method shown in Equation (\ref{eq11}) to calculate the GPII for all solutions in the set $P_{all}$. Next, it performs a global performance impairment ranking from smallest to largest for all solutions in $P_{all}$ based on GPII. After that, the algorithm obtains the complete pruning scheme $S_{all}$ based on the $P_{sort}$ and the pre-set pruning rate $P$.

\begin{algorithm}[h]
\small
\caption{The multiple sub-networks joint pruning method based on EMO.}
\label{algorithm1}
\KwIn{Sub-netwoks $Net_1, Net_2, \cdots, Net_M$. Unpruned network $Net_{all}$. Pruning rate $P$.}
\KwOut{The complete pruning scheme $S_{all}$.}
Perform evolutionary multi-objective pruning on all sub-network $Net_1, Net_2, \cdots, Net_M$ independently\;
$P_{all} \leftarrow$ get the populations on the Pareto Front of every sub-network optimization\;
$I \leftarrow$ Calculate the GPII of every solution on $P_{all}$ with the Equation (\ref{eq11})\;
$P_{sort} \leftarrow$ Rank the solutions in $P_{all}$ according to $I$ from smallest to largest (GPIR)\;
$S_1, S_2, ..., S_M \leftarrow$ Initialize all sub-network pruning schemes with the unpruned network $Net_{all}$\;

\While{ $P_{all}$ is not $\emptyset$}{
    $P_{select} \leftarrow$ find the population with smallest GPII in $P_{sort}$\;
    $Net_i \leftarrow$ Find the sub-network to which the $P_{select}$ belongs\;
    $S_i \leftarrow$ Set the pruning scheme used by $P_{select}$ for $Net_i$\;
    Build complete pruning network structure $S$ with $S_1, S_2, \cdots, S_M$\;
    $P_{cur} \leftarrow$ Calculate the pruning rate on the parameter number of the $Net_p$\;
    \If{$P_{cur} \geq P$} {
        \textbf{Return} the complete pruning scheme $S_{all}$.
    }
    $P_{all} = P_{all} - P_{select}$\;
}
\textbf{Return} the complete pruning scheme $S_{all}$.
\end{algorithm}

It is an iterative process from GPIR to a complete pruning scheme. The algorithm first sets the pruning schemes on all sub-networks to the unpruned structure. The pruning schemes are then updated according to the structure selected on the GPIR from smallest to largest. 
This allows the pruning structure of the complete network to be continuously updated based on the GPIR. The new complete pruning scheme with a bigger pruning rate is obtained from a new pruned sub-network structure taken out of the ranking. The final pruned network structure is obtained according to a pre-set pruning rate. In addition, if the user needs a new pruning scheme with a different pruning rate, he only needs to perform the steps from the global performance impairment ranking to the
final complete network pruning scheme from scratch instead of the whole algorithm.

\section{Experimental Settings and Results Analysis}
In this section, the paper describes the experimental settings and the analysis. First, the datasets used in this paper and algorithm performance evaluation criteria are presented. Then, the experimental parameter settings associated with the proposed algorithm are presented. Then, these experimental results and analysis are provided in detail. Finally, extensive performance analysis experiments are conducted
for understanding the proposed algorithm EMO-DIR better.

\subsection{Benchmark Datasets and Performance Evaluation Criteria}
Four image classification datasets are used in this paper. They are CIFAR10 \cite{krizhevsky_learning_2009}, CIFAR100 \cite{krizhevsky_learning_2009}, and ImageNet-100 \cite{tsai_learning_2022}, and ImageNet-1K \cite{deng_imagenet_2009}. CIFAR10 and CIFAR100 contain 50,000 training and 10,000 testing samples with $32 \times 32$ color images. The CIFAR10 dataset contains 10 categories of samples, while the CIFAR100 dataset contains 100 categories of samples. The ImageNet-100/1K dataset contains 100/1000 categories of color images, and these images vary in size. This dataset contains 127K/1.28M training samples and 5K/50K testing samples.

For conveniently comparing the performance of the proposed algorithm with other network pruning algorithms, three performance evaluation criteria are used, namely the network's prediction error on the testing set, the pruning rate of floating point operations, and the pruning rate of the parameter number. The pruning rate for floating-point operations and the one for parameter number are given in equations (\ref{eq12}) and (\ref{eq13}).

\begin{equation}
\label{eq12}
P_{Flops} = 1 - \frac{Flops(Model_{pruned})}{Flops(Model_{unpruned})}
\end{equation}

\begin{equation}
\label{eq13}
P_{Parameters} = 1 - \frac{Param(Model_{pruned})}{Param(Model_{unpruned})}
\end{equation}
where $Flops(\cdot)$ denotes the calculation function for the amount of floating-point operations and $Param(\cdot)$ denotes the one for the parameter number. $Model_{unpruned}$ denotes the unpruned network model and $Model_{pruned}$ denotes the pruned network model.

\subsection{Experimental Parameter Settings}
This paper uses two kinds of parameters. One is used for the EMO algorithm. The evolutionary algorithm used is NSGA-II \cite{deb_fast_2002}. 
The crossover and mutation operators are the simulated binary crossover and the polynomial mutation operator. 
More detailed settings for this kind of parameter are shown in Table \ref{table1}. The experiments performed in this paper are carried out on the ResNet \cite{he_deep_2016} and VGG \cite{simonyan_very_2015} networks, which involve four network structures, ResNet56, ResNet110, ResNet50, and VGG16. To simplify the division of sub-networks, the network layers dealing with the same spatial resolution features are divided into the same sub-network. Considering that the number of sub-network structures divided into ResNet56 and ResNet110 networks differs from that of ResNet50 and VGG16, their parameter settings are also different. Dividing more sub-networks can reduce the algorithm's optimization complexity and reach a larger combinatorial search space with the same parameter settings. Therefore, the experiments with 4 sub-networks set smaller population sizes and iterations than the ones with 3 sub-networks.

\begin{table}[h]
	\setlength{\abovecaptionskip}{0.cm}
	\setlength{\belowcaptionskip}{0.cm}
	\footnotesize
	\centering
	\renewcommand{\arraystretch}{1.2}
	\caption{
		\centering{Parameter settings related to evolutionary algorithm.}}
	\label{table1}
	\setlength{\tabcolsep}{1.5mm}
	\begin{tabular}{|c|c|c|c|c|}
		\hline
		Network Structure                      & ResNet56                                      & ResNet110    & ResNet50   & VGG16 \\ \hline
		Sub-network Number                & \multicolumn{2}{c|}{3}                  & \multicolumn{2}{c|}{4}          \\ \hline
		Population Size                           & \multicolumn{2}{c|}{40}               & \multicolumn{2}{c|}{20}         \\ \hline
		New Offspring Number              & \multicolumn{2}{c|}{20}               & \multicolumn{2}{c|}{10}         \\ \hline
		Iterations                                      & \multicolumn{2}{c|}{7}                 & \multicolumn{2}{c|}{5}          \\ \hline
		Combinatorial Space & \multicolumn{2}{c|}{5,832,000} & \multicolumn{2}{c|}{24,010,000} \\ \hline
	\end{tabular}
\end{table}

The other is used for the training of the neural network. This paper uses different parameter settings for different datasets, and the detailed settings are shown in Table \ref{table2}. 

\vspace{-2mm}
\begin{table}[!h]
\setlength{\abovecaptionskip}{2.mm}
\setlength{\belowcaptionskip}{0.cm}
\footnotesize
\centering
\renewcommand{\arraystretch}{1.1}
\caption{
\centering{Parameter settings related to network training.}}
\label{table2}
\setlength{\tabcolsep}{1.3mm}
\begin{tabular}{|c|c|c|c|}
\hline
DataSet                 & CIFAR10 & CIFAR100            & ImageNet-100         \\ \hline
full-net Epochs & 300     & 240                         & 100                  \\ \hline
sub-net Epochs  & 100     & 100                         & 50                   \\ \hline
Optimizer                 & \multicolumn{3}{c|}{SGD}                           \\ \hline
Initial LR                & \multicolumn{2}{c|}{0.1}    & 0.05                    \\ \hline
Batch Size                & \multicolumn{3}{c|}{64}                    \\ \hline
Weight decay              & \multicolumn{2}{c|}{0.0005} & 0.0001               \\ \hline
LR scheduler    & Cosine  & MStep {[}150,180,210{]}     & MStep {[}30,60,90{]} \\ \hline
\end{tabular}
\end{table}

The sub-network contains fewer parameters than the full-network structure, and thus it converges faster. In experiments, it converges well in just one-half or even one-third of the training epochs. In addition, all  experiments are implemented in the Pytorch \cite{paszke_pytorch_2019} deep learning library.

\subsection{Experimental Results and Analysis on The Benchmark Datasets}
In order to verify the performance of the EMO-DIR on different datasets, performance tests are conducted on four datasets in this paper. The following part of this subsection will detail the experimental results and the corresponding analysis of these datasets.

\vspace{2mm}
\subsubsection{Experimental Results and Analysis on the CIFAR10 Dataset}
~\par
\vspace{2mm}

The performance of the EMO-DIR was tested on a medium-sized image classification dataset CIFAR10. The neural networks used in the experiments are the ResNet56, ResNet110 and VGG16 networks. The experimental results are shown in Tables \ref{table3}, \ref{table4} and \ref{table4+}, which show the test error (Err), the number of floating point operations (FLOPs), the pruning rate of floating point operations (PR-F), the parameter number (Param), and the pruning rate of the parameter number (PR-P) on the pruned neural networks obtained from different pruning algorithms.

\vspace{-2mm}
\begin{table}[h]
\setlength{\abovecaptionskip}{2.mm}
\setlength{\belowcaptionskip}{0.cm}
\footnotesize
\centering
\renewcommand{\arraystretch}{1.1}
\caption{
\centering{Pruning results of the EMO-DIR and the comparison algorithm on CIFAR10 for ResNet56.}}
\label{table3}
\setlength{\tabcolsep}{1.8mm}
\begin{tabular}{cccccc}
\hline
Method              & Err (\%) & FLOPs & PR-F (\%) & Param & PR-P (\%) \\ \hline
$L_1$-Net \cite{li_pruning_2017}  & 6.94  & 90.9M & 27.6    & 0.73M      & 14.1    \\
CCP-AC \cite{peng_collaborative_2019}  & 6.58  & 59.2M   & 52.8       & -   & - \\
SFR \cite{he_soft_2018}        & 6.65  & 59.4M & 52.7    & -          & -       \\
CD \cite{ayinde_building_2018} & 6.88  & 90.7M & 27.9    & 0.65M      & 23.5    \\
NISP \cite{yu_nisp_2018}       & 6.95  & 70.5M & 43.8    & 0.48M      & 43.5    \\
FPGM \cite{he_filter_2019}     & 6.51  & 59.4M & 52.6    & -          & -       \\
GAL \cite{lin_towards_2019}    & 7.02  & 78.0M & 37.6    & 0.75M      & 11.8    \\
HRank \cite{lin_hrank_2020}    & 6.83  & 62.7M & 50      & 0.49M      & 42.4    \\
ASFP \cite{he_asymptotic_2020} & 6.65  & 59.0M & 52.6    & -          & -       \\
LRMF \cite{zhang_filter_2021}  & 6.71  & 59.4M & 52.6    & -          & -       \\
Niu et al. \cite{niu_exploiting_2022}  & 6.52  & 63.4M   & 49.5       & 0.47M   & 44.7 \\
EMO-DIR  & \textbf{6.29}  & \textbf{49.5M} & \textbf{60.6}    & \textbf{0.37M}      & \textbf{57.0}   \\ \hline
\end{tabular}
\end{table}

Table \ref{table3} depicts the pruning results of the EMO-DIR and the comparison algorithm on the ResNet56. The first column of the table presents the pruning algorithms used, and the second column presents the prediction error of the pruned network obtained by the corresponding algorithm. The third and fourth columns present the pruned network's floating point operation number and the corresponding pruning rate. The fifth and sixth columns show the parameter number and the corresponding pruning rate of the pruned network. The experimental results present that the EMO-DIR performs better on this pruning task. The pruning algorithm proposed by Niu et al. reduces 49.5\% of the floating-point operation number and 44.7\% of the parameter number with an error of 6.52\%. In contrast, the proposed algorithm reduced the pruning rate of the floating-point operation number and the parameter number by 60.6\% and 57.0\% with an error of 6.29\%. This indicates the effectiveness of the EMO-DIR.

\vspace{-2mm}
\begin{table}[h]
\setlength{\abovecaptionskip}{2.mm}
\setlength{\belowcaptionskip}{0.cm}
\footnotesize
\centering
\renewcommand{\arraystretch}{1.1}
\caption{
\centering{Pruning results of the EMO-DIR and the comparison algorithm on CIFAR10 for ResNet110.}}
\label{table4}
\setlength{\tabcolsep}{1.8mm}
\begin{tabular}{cccccc}
\hline
Method            & Err (\%) & FLOPs  & PR-F (\%) & Param & PR-P (\%) \\ \hline
$L_1$-Net \cite{li_pruning_2017}    & 6.45  & 213M   & 15.8    & 1.68M & 2.3     \\
SFR \cite{he_soft_2018}          & 6.14  & 150M   & 40.7    & -     & -       \\
CD \cite{ayinde_building_2018}   & 6.73  & 154M   & 39.1    & 1.13M & 34.3    \\
NISP \cite{yu_nisp_2018}         & 6.61  & 143M   & 43.5    & 0.98M & 43.1    \\
FPGM \cite{he_filter_2019}       & 6.32  & 121M   & 52.2    & -     & -       \\
GAL \cite{lin_towards_2019}      & 7.45  & 130M   & 48.5    & 0.95M & 44.8    \\
HRank \cite{lin_hrank_2020}      & 6.64  & 105.7M & 58.2    & 0.70M & 59.2    \\
CAC \cite{chen_dynamical_2021}   & 6.46  & 123.5M & 51.15   & 0.82M & 52.31   \\
ASFP \cite{he_asymptotic_2020}   & 6.63  & 121M   & 52.3    & -     & -       \\
LRMF \cite{zhang_filter_2021}    & 6.39  & 94.0M  & 62.8    & -     & -       \\
Niu et al. \cite{niu_exploiting_2022} & 6.22  & 126M   & 50.2    & 0.78M & 54.7    \\
EMO-DIR  & \textbf{6.04}  & \textbf{72.2M}  & \textbf{71.5}   & \textbf{0.55M} & \textbf{68.0}   \\ \hline
\end{tabular}
\end{table}

Table \ref{table4} shows the pruning performance of the proposed algorithm and the comparison algorithm on the ResNet110, which has the same structure as Table \ref{table3}. The ResNet110 has more than twice the parameter and floating-point operation numbers than the ResNet56, making this pruning task more difficult. Comparing Table \ref{table3} with Table \ref{table4}, we could find that the pruned network obtained on the ResNet110 contains a larger parameter scale and more floating-point operation numbers under similar errors. For example, the ASFP obtained pruning networks with an error of 6.65\% and 6.63\% on these two networks, respectively, while the pruning network on ResNet110 had twice the parameter number as the pruning network on ResNet56. However, the proposed algorithm still performs well for this pruning task. The state-of-the-art pruning algorithm proposed by Niu et al. \cite{niu_exploiting_2022} achieves 50.2\% and 54.7\% pruning rates for floating point operations and parameters under an error of 6.22\%. The proposed algorithm achieves 71.5\% and 68.0\% pruning rates for floating point operations and parameters with fewer error. This indicates the effectiveness and efficiency of the EMO-DIR in complex pruning tasks.

This paper also tests the pruning performance of the proposed algorithm on the VGG network structure, which is used to test the generalization performance of the proposed algorithm on different network structures. The experimental results are shown in Table \ref{table4+}, which has the same structure as Table \ref{table3}. The network structure used is VGG16.

\vspace{-2mm}
\begin{table}[h]
\setlength{\abovecaptionskip}{2.mm}
\setlength{\belowcaptionskip}{0.cm}
\footnotesize
\centering
\renewcommand{\arraystretch}{1.1}
\caption{
\centering{Pruning results of the EMO-DIR and the comparison algorithm on CIFAR10 for VGG16.}}
\label{table4+}
\setlength{\tabcolsep}{1.8mm}
\begin{tabular}{cccccc}
\hline
Method  & Err (\%)  & FLOPs & PR-F (\%)  & Param & PR-P (\%)  \\ \hline
$L_1$-Net \cite{li_pruning_2017}     & 6.60 & 206M  & 34.2 & 5.4M  & 64.0 \\
CD \cite{ayinde_building_2018}     & 6.33 & 186M  & 40.5 & 3.23M & 78.1 \\
FPGM \cite{he_filter_2019}        & 6.46 & 200M  & 35.9 & -     & -    \\
GAL \cite{lin_towards_2019}        & 9.22 & 172M  & 45.2 & 2.67M & 82.2 \\
Niu et al. \cite{niu_exploiting_2022}  & 6.27 & 138M  & 56.0 & \textbf{2.31M} & \textbf{84.6} \\
EMO-DIR       & \textbf{6.24} & \textbf{96M}   & \textbf{69.4} & 2.86M & 80.6 \\ \hline
\end{tabular}
\end{table}

From the experimental results, compared to the state-of-the-art pruning algorithm proposed by Niu et al., the proposed algorithm obtains a higher pruning rate of FLOPs with similar error and parametric pruning. Their FLOPs is 138M, while the proposed algorithm is 96M.

\vspace{2mm}
\subsubsection{Experimental Results and Analysis on the CIFAR100 Dataset}
~\par
\vspace{2mm}

This paper also tests the EMO-DIR on the CIFAR100 dataset, which has the same training and testing sample numbers as CIFAR10, while CIFAR100 has ten times more categories than CIFAR10. So it is more challenging to classify this dataset. This subsection conducts pruning experiments on the ResNet56 and ResNet110, and 
Tables \ref{table5} and \ref{table6} describe
the pruning results of the EMO-DIR and compared algorithms.

\vspace{-2mm}
\begin{table}[h]
\setlength{\abovecaptionskip}{2.mm}
\setlength{\belowcaptionskip}{0.cm}
\footnotesize
\centering
\renewcommand{\arraystretch}{1.1}
\caption{
\centering{Pruning results of the EMO-DIR and the comparison algorithm on CIFAR100 for ResNet56.}}
\label{table5}
\setlength{\tabcolsep}{2.0mm}
\begin{tabular}{cccccc}
\hline
Method                                              & Err (\%)  & FLOPs  & PR-F (\%) & Param & PR-P (\%) \\ \hline
LCCL \cite{dong_more_2017}         & 31.63     & 76.3M  & 39.3          & -           & -       \\
SFP \cite{he_soft_2018}                  & 31.21      & 59.4M  & 52.6          & -          & -       \\
FPGM \cite{he_filter_2019}             & 30.34     & 59.4M  & 52.6          & -          & -       \\
TAS \cite{dong_network_2019}      & 27.75     & 61.2M   & 51.3           & -          & -       \\
LFPC \cite{he_learning_2020}        & 29.17      & 60.8M  & 51.6          & -           & -       \\
CAC \cite{chen_dynamical_2021}  & 30.22     & 61.4M   & 51.11         & 0.46M & 45.51   \\
EMO-DIR & \textbf{27.54} & \textbf{52.73M} & \textbf{58.07}   & \textbf{0.39M} & \textbf{54.45}   \\ \hline
\end{tabular}
\end{table}

Table \ref{table5} has the same structure as Table \ref{table3}. 
The experimental results demonstrate that the EMO-DIR still performs well on the CIFAR100 dataset.
The state-of-the-art network pruning algorithm TAS obtains a pruning rate of 51.3\% for the parameter number under an error of 27.75\%. In contrast, the proposed algorithm achieves an error of 27.62\%, and its pruning rate of floating-point operations is 6.47 points higher than TAS. This indicates the effectiveness of the EMO-DIR on the CIFAR100 dataset.

\vspace{-2mm}
\begin{table}[h]
\setlength{\abovecaptionskip}{2.mm}
\setlength{\belowcaptionskip}{0.cm}
\footnotesize
\centering
\renewcommand{\arraystretch}{1.1}
\caption{
\centering{Pruning results of the EMO-DIR and the comparison algorithm on CIFAR100 for ResNet110.}}
\label{table6}
\setlength{\tabcolsep}{1.8mm}
\begin{tabular}{cccccc}
\hline
Method              & Err (\%) & FLOPs  & PR-F (\%)    & Param & PR-P (\%)    \\ \hline
Niu et. al \cite{niu_exploiting_2022} & 25.1  & 184M   & 27.5  & 1.29M & 25.1  \\
LCCL \cite{dong_more_2017}       & 29.22 & 173M   & 31.3  & -     & -     \\
SFP \cite{he_soft_2018}        & 28.72 & 121M   & 52.3  & -     & -     \\
FPGM \cite{he_filter_2019}       & 27.45 & 121M   & 52.3  & -     & -     \\
TAS \cite{dong_network_2019}        & 26.84 & 120M   & 52.6  & -     & -     \\
CAC \cite{chen_dynamical_2021}        & 28.19 & 123.5M & 51.18 & 0.93M & 45.77 \\
EMO-DIR & \textbf{24.96} & \textbf{108.6M} & \textbf{57.12} & \textbf{0.83M} & \textbf{52.09} \\ \hline
\end{tabular}
\end{table}

Table \ref{table6} has the same structure as Table \ref{table3}. 
The experimental results demonstrate that the EMO-DIR still performs well on the pruning task of the complex ResNet110.
For the advanced pruning algorithm TAS, the proposed algorithm obtained lower error with similar pruning rates of floating point operations, and the error of the proposed algorithm was 1.88 points lower than TAS. In addition, the proposed algorithm obtains a higher pruning rate with similar error compared to the algorithm proposed by Niu et al. Its pruning rate of floating point operations and parameters reaches is more than twice that of the algorithm proposed by Niu et al. This indicates the effectiveness and efficiency of the EMO-DIR.

\vspace{2mm}
\subsubsection{Experimental Results and Analysis on the ImageNet-100 and ImageNet-1K Dataset}
~\par
\vspace{2mm}
This paper also conducts evolutionary multi-objective pruning experiments on the ImageNet-100 dataset. This dataset has more data samples, each with a higher spatial resolution. The pruning task on this dataset is more complicated. Then, the pruned network structure was migrated to the larger dataset, ImageNet-1K, which is used to test the generalization performance of the pruned structure from the proposed algorithm on the larger dataset. We have only modified the output class number of the last fully connected layer to accommodate the new dataset. Table \ref{table7} records the corresponding experiment results, where the network structure used in this experiment is ResNet50.

\vspace{-2mm}
\begin{table}[h]
\setlength{\abovecaptionskip}{2.mm}
\setlength{\belowcaptionskip}{0.cm}
\footnotesize
\centering
\renewcommand{\arraystretch}{1.1}
\caption{
\centering{Pruning results of the EMO-DIR and the comparison algorithm on ImageNet-100/1K for ResNet50.}}
\label{table7}
\setlength{\tabcolsep}{1.2mm}
\begin{tabular}{cccccc}
\hline
\multicolumn{6}{c}{ImageNet-100} \\ \hline
Method  	 & Err Top1/Top5		& FLOPs & PR-F (\%) & Param  & PR-P (\%) \\ \hline
Baseline 	  & 15.70 / 3.88    & 4.09G & 0.00          & 25.56M & 0.00      \\
EMO-DIR  & 15.90 / 3.90    & 2.10G & 48.74         & 12.94M & 45.43     \\	\hline \hline
\multicolumn{6}{c}{ImageNet-1K} \\ \hline
Method           		                                          & Err Top1/Top5        & FLOPs & PR-F (\%)    & Param   & PR-P (\%) \\ \hline

He et al. \cite{he_channel_2017}       		      & 27.70 / 9.20 		  & 2.73G & 33.3      		 & -      		 & -               \\
GAL-0.5 \cite{lin_towards_2019}          		& 28.05 / 9.06 		& 2.33G & 43.0      		& 21.20M & 16.9        \\
SSS-26  \cite{huang_data_driven_2018}    & 28.18 / 9.21 		& 2.33G & 43.0      		& 15.60M & 38.8       \\
EMO-DIR 			                                          & \textbf{27.31} / \textbf{8.85} 		  & \textbf{2.10G} & \textbf{48.7}     	  	 & \textbf{14.78M} & \textbf{42.2}        \\  \hline
\end{tabular}
\end{table}

Table \ref{table7} demonstrates that the proposed algorithm EMO-DIR still performs well on the ImageNet-100 dataset. It achieves a pruning rate of 48.74\% for the floating point operation and 45.43\% for the parameter number, with a drop of 0.20 and 0.02 points in Top-1 and Top-5 performance. This indicates the effectiveness and efficiency of the EMO-DIR on the model pruning task for large-scale datasets.
In experiments testing the generalization performance of the pruned structure from ImageNet-100 on ImageNet-1K, the pruned network obtained by the proposed algorithm outperforms the other three pruning algorithms. The proposed algorithm obtains higher pruning rates for the floating-point operation and the parameter number with lower Top-1 and Top-5 performance losses. This shows that the network structure pruned by the proposed algorithm EMO-DIR could perform well on the similar dataset.

\subsection{Performance Analysis of The Proposed Algorithm}
From the above experiments, the EMO-DIR performs well on several datasets. In this section, we perform more experiments to further analyze the performance of the EMO-DIR.

\vspace{2.5mm}
\subsubsection{Performance Comparison Between The Whole and The Divide-and-conquer Optimization}
~\par
\vspace{2.5mm}

The divide-and-conquer idea is the key to the proposed algorithm, which decomposes the complex evolutionary multi-objective pruning task on the whole network into simple evolutionary multi-objective pruning tasks on several smaller sub-networks. On the one hand, this idea can reduce the optimization complexity of the evolutionary multi-objective algorithm. On the other hand, it can reduce the computational complexity of network performance verification. In order to understand the proposed divide-and-conquer evolutionary multi-objective network pruning method further, this paper designs a whole optimization algorithm EMO-W and conducts corresponding analytical experiments. EMO-W uses the whole optimization idea instead of the divide-and-conquer in the EMO-DIR algorithm, which encodes and optimizes the whole network as a sub-network. In order to allow the network to converge sufficiently, it uses full training epochs when verifying the performance of the pruned network. The rest of the settings are the same as for EMO-W. The network used was ResNet56; the dataset was CIFAR100. Table \ref{table8} describes the experimental results. It compares the prediction error, the floating-point operation number, the pruning rate of the floating-point operations, the parameter number, the pruning rate of the parameters, and the running time from these two methods.

\vspace{-2mm}
\begin{table}[h]
\setlength{\abovecaptionskip}{2.mm}
\setlength{\belowcaptionskip}{0.cm}
\footnotesize
\centering
\renewcommand{\arraystretch}{1.1}
\caption{
\centering{Pruning results of the EMO-DIR and EMO-W on the CIFAR100 for the ResNet56.}}
\label{table8}
\setlength{\tabcolsep}{1.3mm}
\begin{tabular}{ccccccc}
\hline
Method        & Error (\%)  & FLOPs  & PR (\%) & Param & PR (\%) & GPU Day \\ \hline
EMO-W         & 27.44        & 65.81M & 47.66    & 0.46M & 46.63   & 4.9     \\
EMO-DIR            & 27.54        & 52.73M & 58.07    & 0.39M & 54.45    & 2.3 \\ \hline
\end{tabular}
\end{table}

Table \ref{table8} shows that the EMO-DIR obtains better pruning performance. The pruning network designed by the EMO-DIR obtains a higher pruning rate on the floating-point operation and the parameter number with similar performance. Its pruning rate on the floating-point operation is 10.41 points higher than the EMO-W, and the parameter number is 7.82 points higher than the EMO-W. In addition, The EMO-DIR consumes a lower running time, and the time used is lower than half of the EMO-W, thanks to the fast convergence speed of the sub-network.

\begin{figure}[h]
\centering
\includegraphics[scale=0.6]{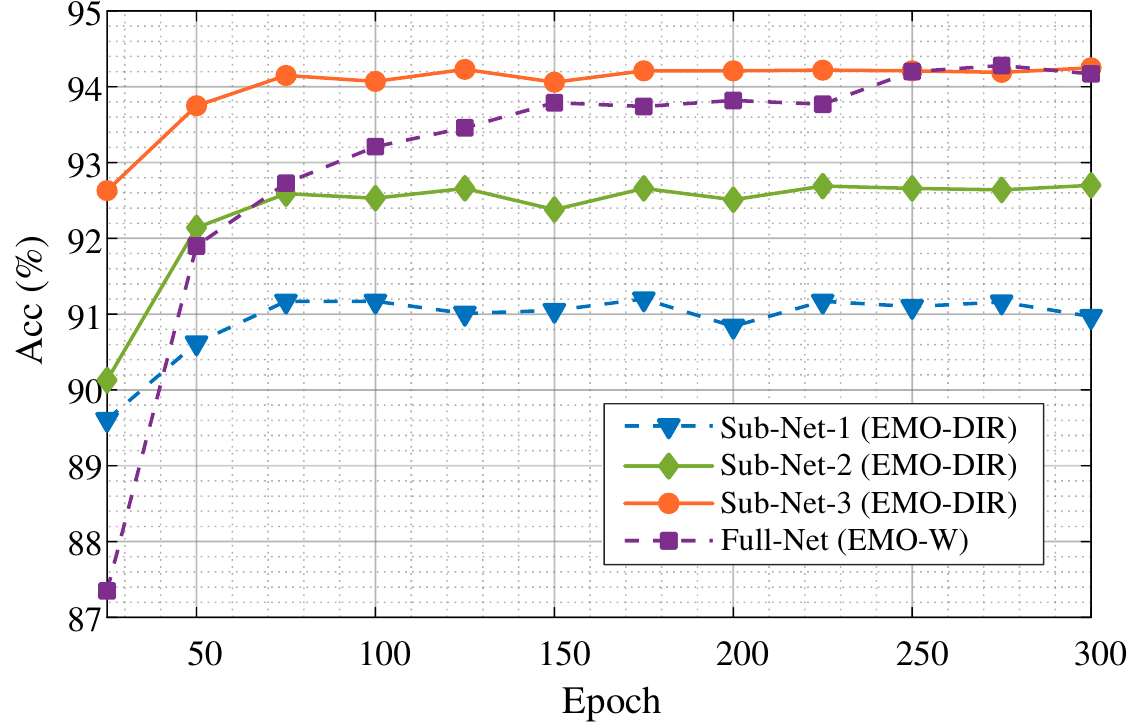}
\caption{The convergence of network structures in EMO-DIR and EMO-W with different training epochs.}
\label{fig7}
\end{figure}

To further analyze the network convergence under the two methods, the network training process under the two methods is also compared in Fig. \ref{fig7}. The networks are the unpruned network structures. Compared with the pruned network, the complexity of the unpruned network structure is higher, which is more conducive to comparing the convergence of the network structure under the two methods. In this experiment, the complete network and the sub-network were trained at different training epochs, and their convergence accuracy was recorded.

\vspace{1mm}
\vspace{1mm}

Fig. \ref{fig7} shows that the whole network Full-Net converges more slowly than the sub-network Sub-Net-1, Sub-Net-2, and Sub-Net-3, and Full-Net requires a longer training epoch to converge to a stable accuracy. In the experiments, the Full-Net performance keeps growing as the training epoch grows and only stabilizes when it exceeds 250 epoch. In contrast, the accuracy of the other three sub-networks stabilizes when their training epoch exceeds 75. The training epoch for convergence to stable accuracy is less than one-third of that for the whole network structure. In addition, These three sub-networks converge at different accuracies. The sub-network closer to the network input has a lower accuracy, and the sub-network closer to the output has a higher accuracy. This is because different sub-networks focus on different levels of feature extraction. The closer to the network's output, the more advanced the features produced by the sub-network, and the less complicated it is for a simple feature detector to extract the correct class.

From the above experiments, it is worth mentioning that the idea based on divide-and-conquer can reduce the difficulty of the evolutionary multi-objective pruning algorithm in complex networks. On the one hand, this idea reduces the optimization space of the algorithm, reducing its difficulty. On the other hand, it reduces the parameter number of the network to be verified, accelerating the network's convergence and reducing the algorithm's computational complexity. This enables the proposed algorithm to obtain better network pruning with less computational complexity.

\vspace{2.5mm}
\subsubsection{The Effect of Cross-network Constraints on Cooperation between Sub-networks}
~\par
\vspace{2.5mm}

The proposed sub-network training method based on the cross-network constraints method can constrain the previous sub-network's output feature and the following one's input feature to enhance the relationship among these sub-networks optimized independently, improving the overall performance of the optimized network. In order to verify the impact of this method on the EMO-DIR, an algorithm EMO-DIR-nc is designed that does not use cross-network constraints. It removes the feature constraints term in Equation (\ref{eq9}) and only trains the sub-networks with the cross-entropy loss. The dataset used is CIFAR100, the network is ResNet56, and Table \ref{table9} describes the experimental results.

\begin{table}[h]
\setlength{\abovecaptionskip}{2.mm}
\setlength{\belowcaptionskip}{0.cm}
\footnotesize
\centering
\renewcommand{\arraystretch}{1.2}
\caption{
\centering{Pruning results of the EMO-DIR and EMO-DIR-nc for the ResNet56 on the CIFAR100.}}
\label{table9}
\setlength{\tabcolsep}{2.mm}
\begin{tabular}{cccccc}
\hline
Method & Error (\%) & FLOPs  & PR (\%) & Param & PR (\%) \\ \hline
EMO-DIR-nc  & 27.43 & 54.74M & 56.47   & 0.44M & 48.50   \\
EMO-DIR   & 27.54        & 52.73M & 58.07    & 0.39M & 54.45   \\ \hline
\end{tabular}
\end{table}

The experimental results show that the cross-network feature constraints improve the proposed algorithm's performance. Under similar performance, the proposed algorithm improves the pruning rate on the parameters by approximately 6 points and the pruning rate on the floating-point operations by approximately 1.6 points over the method without cross-network feature constraints. These results demonstrate the excellent performance of the proposed cross-network feature constraint method. 

\begin{figure}[h]
\centering
\includegraphics[scale=0.35]{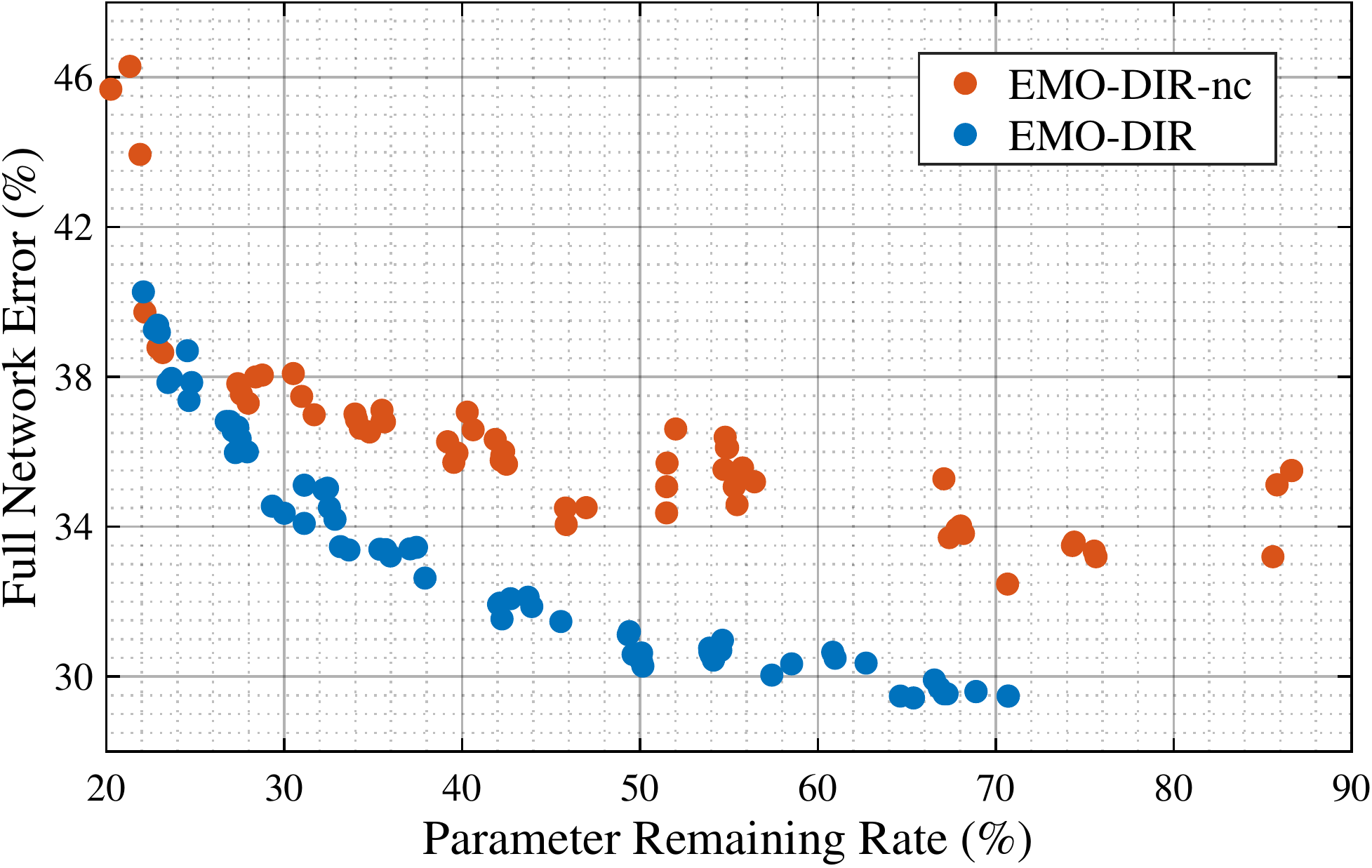}
\caption{Network pruning rates and parameter remaining rates network structures obtained by the EMO-DIR and EMO-DIR-nc through GPIR.}
\label{fig8}
\end{figure}

In addition, to further analyze the effect of the cross-network feature constraint on the relationship among pruned sub-networks, corresponding performance measurements are also performed on all pruned network structures obtained from the global performance impairment ranking in both algorithms. The measurements use the weights of the sub-network obtained during the evolutionary optimization process without additional updates. If these sub-networks trained independently have good collaboration, the complete network performs well on the target task. Furthermore, because the sub-network pruning structures involved in the global performance ranking are all non-dominated solutions on each sub-network pruning task, the network pruning structure constructed by global performance impairment ranking is also a collection of non-dominated solutions if the collaboration among the sub-networks is strong. Because the network structure is only changed one sub-network at a time when constructing the network structure with the global performance impairment index, and both the original structure and the adjusted one are non-dominated solutions on the same sub-network pruning task, the network structure before and after the pruning processes remains on the collection of non-dominated solutions. Therefore, if the collaboration among sub-network structures optimized independently is strong, the sub-structures obtained during optimization can still collaborate well without updating the weights. Even considering the random interference of the network performance verification, the network structure constructed by the global performance impairment ranking is still approximately a collection of non-dominated solutions when using the weights obtained during the evolutionary process. For this reason, this paper conducts the corresponding performance analysis experiments using all the network structures constructed by the global performance impairment ranking in Table \ref{table9}. The error and parameter remaining rates for all network structures are shown in Fig. \ref{fig8}. The weights for all network structures in the experiment were obtained from the training during evolutionary multi-objective optimization without any update.

The experiments show that the network structure constructed by the proposed algorithm is approximately distributed like a Pareto Front. In contrast, the EMO-DIR-nc algorithm, which removes the cross-network constraints, constructs a network structure with a more discrete distribution containing more dominant solutions. Moreover, the EMO-DIR algorithm also performs better at the same pruning rate, which suggests that the cross-network constraints reinforce the relationship among sub-networks and improve the network's overall performance consisting of sub-networks optimized independently.

\vspace{2mm}
\subsubsection{The Relationship Between the Sub-Network Performance and the Feature Changes}
~\par
\vspace{2mm}

The feature detectors can calculate the classification accuracy, measuring the feature processing ability of the sub-networks. However, the change between the features generated by the pruning sub-network and the unpruned one can also measure the drop in the feature processing ability. To this end, an evolutionary multi-objective pruning algorithm based on feature change, EMO-DIR-$L_2$, is designed, which uses the $L_2$ norm between the features generated by the pruning sub-network and the unpruned sub-network as a performance drop measure. The remaining experimental settings are the same as the EMO-DIR. Table \ref{table10} records the experimental results.

\vspace{-2mm}
\begin{table}[h]
\setlength{\abovecaptionskip}{2.mm}
\setlength{\belowcaptionskip}{0.cm}
\footnotesize
\centering
\renewcommand{\arraystretch}{1.1}
\caption{
\centering{Pruning results of the EMO-DIR and EMO-DIR-$L_2$ for the ResNet56 on the CIFAR100.}}
\label{table10}
\setlength{\tabcolsep}{2.mm}
\begin{tabular}{cccccc}
\hline
Method                       & Error (\%)  & FLOPs    & PR (\%) & Param   & PR (\%) \\ \hline
EMO-DIR-$L_2$      & 27.58        & 60.78M  & 51.67   & 0.40M    & 53.08   \\
EMO-DIR                  & 27.54        & 52.73M & 58.07    & 0.39M    & 54.45   \\ \hline

\end{tabular}
\end{table}

The network structure used was ResNet56; the dataset was CIFAR100. The experimental results demonstrate that the EMO-DIR is better than the method using the change between features as the sub-network performance measure. With similar parameter pruning rates and errors, the proposed algorithm has a 6.4 points higher pruning rate for floating point operations than the EMO-DIR-$L_2$. The change between features, while reflecting the drop in the feature processing ability of the pruned sub-network relative to the unpruned one, is not the same as the actual performance drop. Features are not simply collections of elements but also complex interrelationships among elements. The $L_2$ norm can accurately measure the change in the feature elements, but the change in the complex relationships among feature elements is difficult to measure accurately. The feature detector designed for unpruned sub-network can detect and process the features generated by the sub-network and translate the valuable information detected into classification accuracy. Thus, the feature detector provides a better measure of the feature processing capability of the sub-network. 

\begin{figure}[h]
\centering

\includegraphics[scale=0.32]{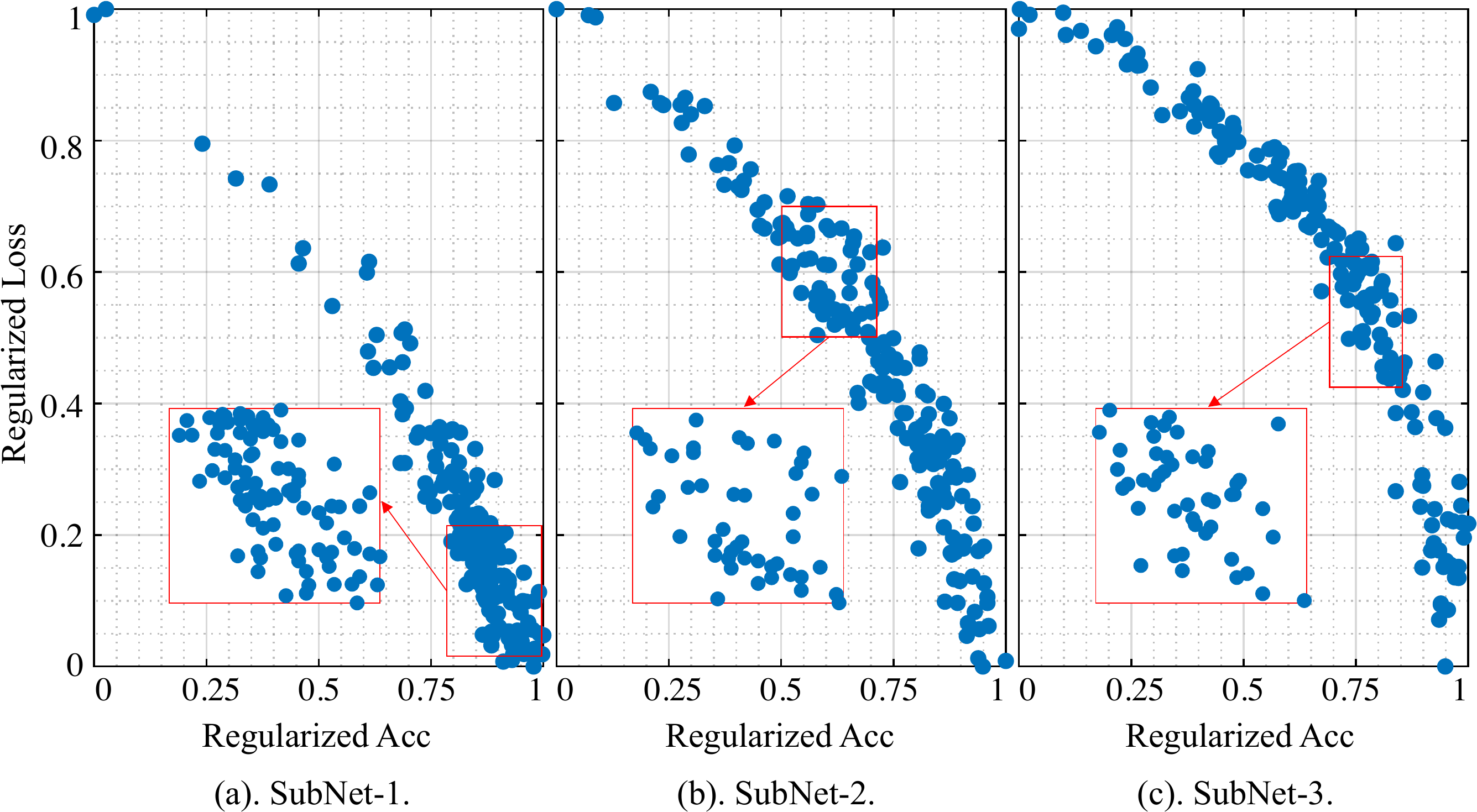}
\caption{Relationship between the feature changes brought by the pruned sub-network and the prediction accuracy of its corresponding feature detector.}
\label{fig9}
\end{figure}

In addition, this paper also compares the change in the feature elements produced by the pruned sub-network with the prediction accuracy produced by the feature detector. 
The change of the feature element (L2 losses) and accuracies are regularized to [0, 1] for a clear comparison. They are all functions of the network pruning structure and change as the pruning scheme changes. Fig. \ref{fig9} depicts the relationship between the regularized losses and the regularized accuracies for all the sub-networks generated during the evolutionary process. 
The three sub-networks in this picture come from the independent evolutionary multi-objective optimization of three sub-networks divided by the same network.

The experimental results show a negative correlation between the feature change and accuracy. This explains why EMO-DIR using accuracy and EMO-DIR-$L_2$ using feature change perform well in the evolutionary multi-objective network pruning task. However, the local zoom in Fig. \ref{fig9} shows that the feature change cannot reflect the change in accuracy well. A reduction in the feature change may lead to a reduction in accuracy. Therefore, a pre-trained feature detector can better guide the evolutionary multi-objective pruning on the sub-networks.

\section{Conclucsion}
EMO network compression algorithms have to face the complex optimization space and the high resource-consuming compression network verification when dealing with the compression task on complex network structures, which limits its application in this field. To this end, this paper proposes 
a multi-objective complex network pruning framework based on divide-and-conquer and global performance impairment ranking
to reduce the space complexity and resource consumption of pruning algorithms based on EMO. Firstly, a divide-and-conquer EMO network pruning method is designed. This method can reduce the optimization difficulty of multi-objective pruning algorithms and the resource consumption of pruning structure verification. In the pruning task on ResNet56 at CIFAR100, compared to the whole network pruning method, this proposed method achieves a 10.41 points improvement in the pruning rate of floating-point operations, a 7.82 points improvement in the pruning rate of parameters and consumed only about half of the running time under a similar performance. This paper then designs a sub-network training method based on cross-network constraints. It enables sub-networks pruned independently to collaborate better by constraining the previous sub-network's output features and the next one's input features, improving the pruned network's overall performance. In an experimental analysis on ResNet56 at CIFAR100, compared to the method without feature constraints, the proposed algorithm brings about a 6 points improvement in the pruning rate of parameters and a 1.6 points improvement in the pruning rate of floating-point operations with similar performance. Finally, this paper designs a multiple sub-networks joint pruning method based on EMO. It combines multi-objective pruning results on multiple sub-networks to design a joint pruning scheme. The effectiveness and efficiency of the EMO-DIR are validated on three datasets of different sizes.

This paper applies the divide-and-conquer idea to evolutionary multi-objective pruning. There are several aspects for improvement. For example, how to further reduce the resource consumption of the algorithm and improve the efficiency of the EMO algorithm. We will continue our research in this area in future work.

\bibliographystyle{IEEEtran}
\bibliography{IEEEabrv,./mybib.bib}



\vspace{-10pt}
\begin{IEEEbiography}[{\includegraphics[width=1in,height=1.25in,clip,keepaspectratio]{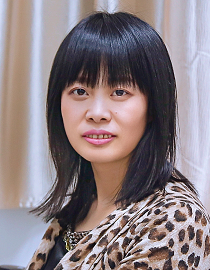}}]{Ronghua Shang} (SM’ 22) received the B.S. degree in information and computation science and the Ph.D. degree in pattern recognition and intelligent systems from Xidian University in 2003 and 2008, respectively. She is currently a professor with Xidian University. Her current research interests include optimization problems, evolutionary computation, image processing, and data mining.\par
\end{IEEEbiography}

\vspace{-10pt}
\begin{IEEEbiography}[{\includegraphics[width=1in,height=1.25in,clip,keepaspectratio]{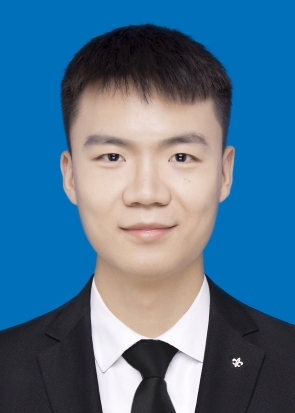}}]{Songling Zhu} received the B.E. degree in School of Electric Power from North China University of Water Resources and Electric Power, Henan, China. Now he is pursuing a Ph.D. degree in School of Artificial Intelligence from Xidian University, Xi’an, China. His current research interests include deep learning, evolutionary deep learning, knowledge distillation, model pruning, model compression.\par
\end{IEEEbiography}

\vspace{-20pt}
\begin{IEEEbiography}[{\includegraphics[width=1.2in,height=1.4in,clip,keepaspectratio]{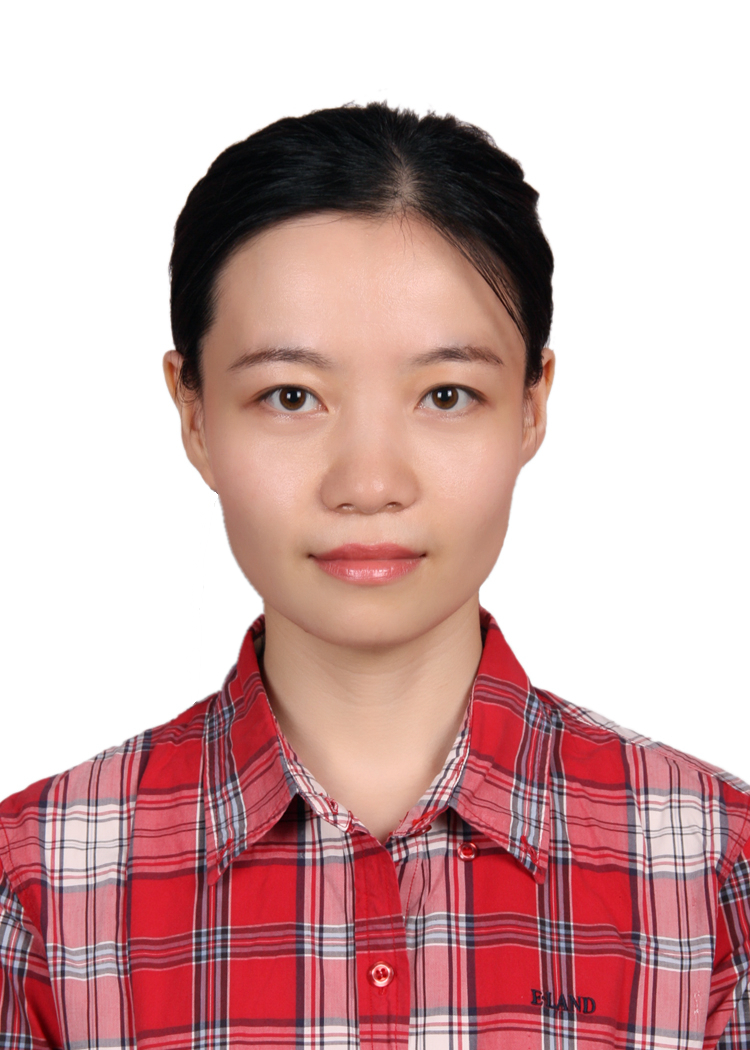}}]{Yinan Wu} received the B.Sc. degree in information and computing science from Northwestern Polytechnical University, Xi'an, China, in 2016. She is currently pursuing a Ph.D. degree with the School of Artificial Intelligence and the Key Laboratory of Intelligent Perception and Image Understanding, Ministry of Education of China, Xidian University, Xi'an, China.
	Her research interests include image/signal processing and analysis of deep learning in Earth Observation.
\end{IEEEbiography}

\vspace{-20pt}
\begin{IEEEbiography}[{\includegraphics[width=1.in,height=1.4in,clip,keepaspectratio]{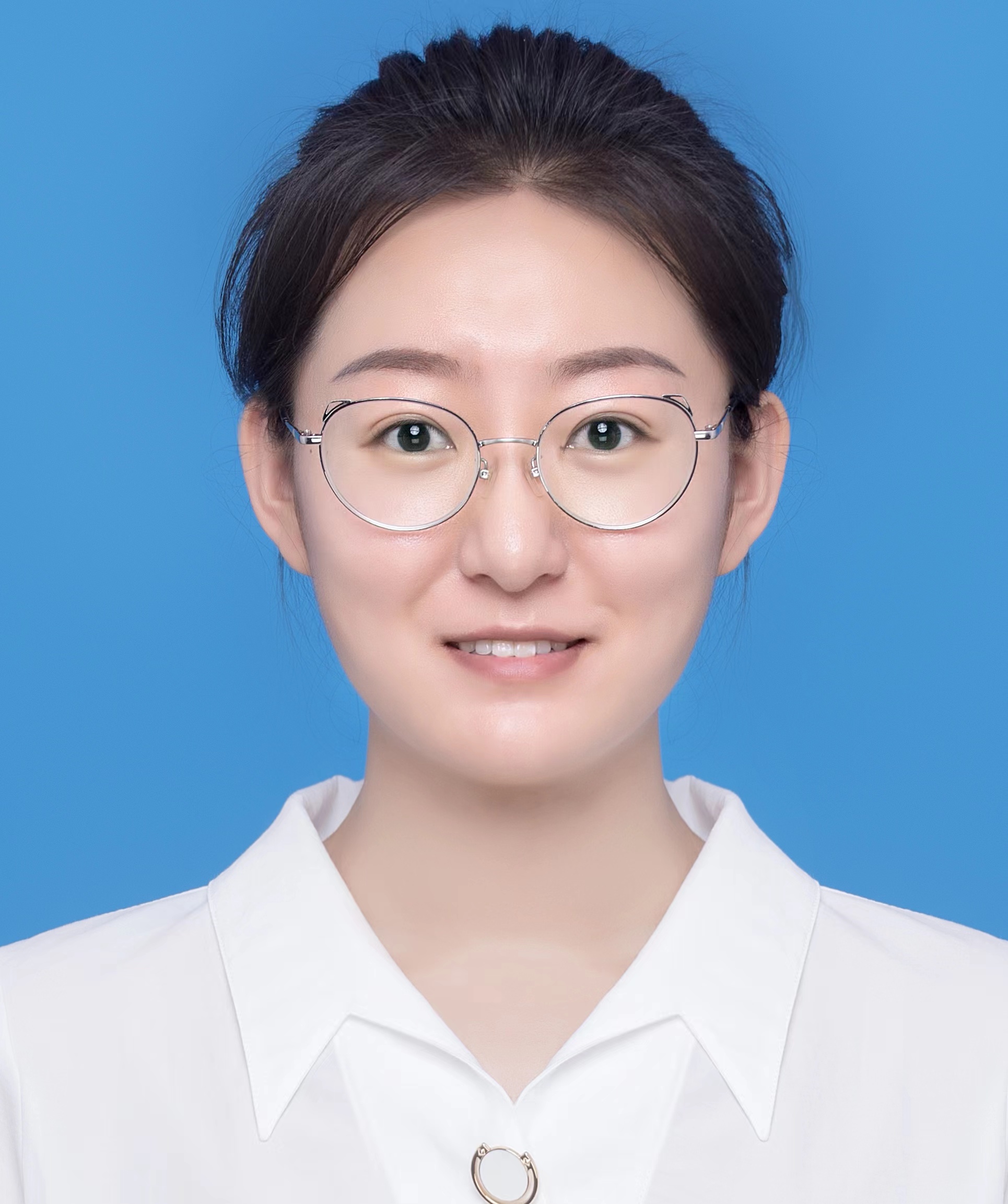}}]{Weitong Zhang} (M’21) received the B.E. degree in Electronic and Information Engineering from Changchun University of Science and Technology, Changchun, China, in 2013, the M.S. degree in Electronics and Communication Engineering, and the Ph.D. degree in Electronic science and technology from Xidian University, Xi’an, China, in 2017 and 2021. She is currently a lecturer with Xidian University. Her current research interests include complex networks and machine learning.
\end{IEEEbiography}

\vspace{-20pt}
\begin{IEEEbiography}[{\includegraphics[width=1in,height=1.25in,clip,keepaspectratio]{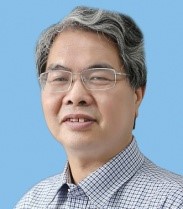}}]{Licheng Jiao} (SM'89-F'17) received the B.S. degree from Shanghai Jiaotong University, Shanghai, China, in 1982, the M.S. and Ph.D. degrees from Xian Jiaotong University, Xian, China, in 1984 and 1990, respectively. From 1990 to 1991, he was a postdoctoral Fellow in the National Key Laboratory for Radar Signal Processing, Xidian University, Xian, China. Since 1992, Dr. Jiao has been a Professor in the School of Electronic Engineering at Xidian University. Currently, he is the Director of the Key Lab of Intelligent Perception and Image Understanding of Ministry of Education of China at Xidian University, Xian, China. Dr. Jiao is a fellow of IEEE, member of IEEE Xian Section Executive Committee and the Chairman of Awards and Recognition Committee, vice board chairperson of Chinese Association of Artificial Intelligence, councilor of Chinese Institute of Electronics, committee member of Chinese Committee of Neural Networks, and expert of Academic Degrees Committee of the State Council. His research interests include image processing, natural computation, machine learning, and intelligent information processing. He has led 40 major scientific research projects, and published more than 20 monographs and a hundred papers in international journals and conferences.\par
\end{IEEEbiography}

\vspace{-20pt}
\begin{IEEEbiography}[{\includegraphics[width=1.2in,height=1.4in,clip,keepaspectratio]{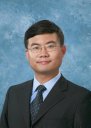}}]{Songhua Xu} is a computer scientist. He received his M.S., M.Phil., and Ph.D. from Yale University, New Haven, CT, USA, all in computer science. His research interests include healthcare informatics, information retrieval, knowledge management and discovery, intelligent web and social media, visual analytics, user interface design, and multimedia.\par
\end{IEEEbiography}

\vfill

\end{document}